\documentclass{article}

\usepackage{arxiv}

\usepackage[utf8]{inputenc} 
\usepackage[T1]{fontenc}    
\usepackage{hyperref}       
\usepackage{url}            
\usepackage{booktabs}       
\usepackage{amsfonts}       
\usepackage{nicefrac}       
\usepackage{microtype}      
\usepackage{lipsum}
\usepackage{graphicx}
\usepackage{subcaption}
\usepackage{longtable}
\usepackage{multirow}
\usepackage{multicol}
\usepackage{ragged2e}
\usepackage{float}
\usepackage{amsmath}
\usepackage{gensymb}
\usepackage{adjustbox}
\usepackage[title]{appendix}
\usepackage[square, numbers]{natbib}
\graphicspath{ {./images/} }

\title{Class-Incremental Learning for Honey Botanical Origin Classification with Hyperspectral Images: A Study with Continual Backpropagation}

\author{
 Guyang Zhang \\
  Department of Electrical, Computer and Software Engineering\\
  University of Auckland\\
  5 Grafton Rd, Auckland Central, Auckland 1010, New Zealand \\
  \texttt{gzha422@aucklanduni.ac.nz} \\
   \And
 Waleed Abdulla \\
  Department of Electrical, Computer and Software Engineering\\
  University of Auckland\\
  5 Grafton Rd, Auckland Central, Auckland 1010, New Zealand \\
  \texttt{w.abdulla@auckland.ac.nz} \\  
}

\begin{document}
\maketitle
\begin{abstract}
Honey is an important commodity in the global market. Honey types of different botanical origins provide diversified flavors and health benefits, thus having different market values. Developing accurate and effective botanical origin-distinguishing techniques is crucial to protect consumers' interests. However, it is impractical to collect all the varieties of honey products at once to train a model for botanical origin differentiation. Therefore, researchers developed class-incremental learning (CIL) techniques to address this challenge. This study examined and compared multiple CIL algorithms on a real-world honey hyperspectral imaging dataset. A novel technique is also proposed to improve the performance of class-incremental learning algorithms by combining with a continual backpropagation (CB) algorithm. The CB method addresses the issue of loss-of-plasticity by reinitializing a proportion of less-used hidden neurons to inject variability into neural networks. Experiments showed that CB improved the performance of most CIL methods by 1-7\%. 
\end{abstract}


\section{Introduction}
\begin{figure}[ht]
		\hspace{-0.3in}
		\includegraphics[scale=0.4]{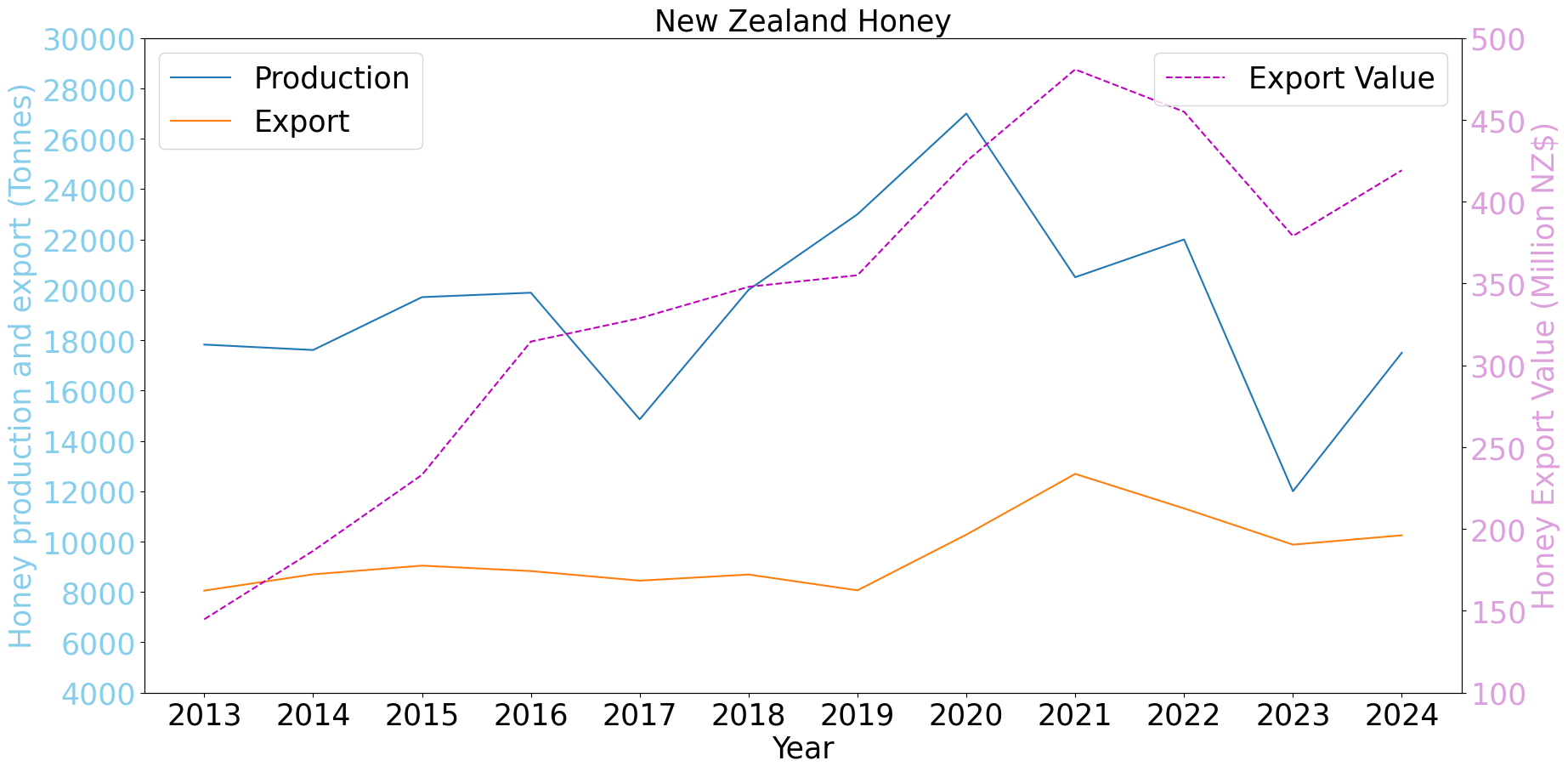}
		\caption{New Zealand honey production and export volumn in tonnes from 2013 to 2024. The dashed line represents the New Zealand honey export value in million NZ\$ \citep{mpinz2024}}
		\label{nzh}
	\end{figure}

	As the only naturally synthesized sweetener that humans can consume directly \citep{codexstan12, eu2001}, honey has become an important commodity in the global market. The growing demand for honey products requires the authentication of honey botanical origins because the various botanical origins of honey provide distinct flavors and health benefits \citep{ULBERTH2016729, SIDDIQUI2017687}. As one of the major honey producers, New Zealand is estimated to export more than 400 million dollars of honey products in 2024, which increased from 23 million export revenue in 2004 \citep{nzmpi2024}. Fig \ref{nzh} displays New Zealand's annual honey production and export volume from 2013 to 2024 with the export value \citep{mpinz2024}. Moreover, as the special botanical origin type in New Zealand, Mānuka honey has a much higher value than other honey types. For instance, the bulk honey price range for light clover honey in New Zealand was 5.00~8.00 NZ\$/kg, while the price range for Mānuka honey was 5.50~180.00 NZ\$/kg \citep{mpinz2024}. Therefore, designing accurate and reliable honey botanical origins authentication detection techniques is essential for the consumers' confidence and the income of honey producers. Researchers have developed numerous techniques to determine honey botanical origins. However, most of them are time-consuming and require tedious processing procedures. In addition, these methods could only process one sample at a time. Given the large amount of honey types, hyperspectral imaging is the only method to process multiple samples simultaneously \citep{Elmasry2010}. 
	
	The hyperspectral images are three-dimensional data cubes that contain hundreds of spectral channels and thousands/even millions, of spatial pixels. The scanned materials' different chemical characteristics and physical structures reflect, absorb, and emit electromagnetic signals with distinctive patterns at different wavelengths \citep{Elmasry2010}. Moreover, the preparation of HSI is non-destructive, fast, and simple, making HSI appropriate for various applications in multiple areas, including assessing sweetpotato quality \citep{AHMED2024108855}, detecting leaf level apple mosaic disease \citep{LIU2024109051}, distinguishing rice seed \citep{GE2024108776}, and detecting wheat disease \citep{XIE2024108571}.
	
	With the advancements in deep learning techniques, an increasing number of classification/regression tasks have utilized deep learning algorithms in food and agricultural areas \citep{BARBEDO2023107920}. However, real-world training data often comes in stream \citep{Lange2022survey}. For instance, it is impossible to collect all different types of honey products at once to train a model to distinguish their botanical origins. The model must be able to learn the critical knowledge from newly obtained honey classes incrementally. The capability of a model to continuously learn over time by incorporating new knowledge while maintaining formerly obtained information is known as continual learning, lifelong learning, or incremental learning. \citep{Zhou2024CILsurvey}. 
	
	Incremental learning, continual learning, or lifelong learning aims to learn the critical knowledge to address new tasks incrementally while maintaining formerly obtained information  \citep{Zhou2024CILsurvey}. In general, incremental learning scenarios can be categorized into three fundamental types: task-incremental learning (TIL), domain-incremental learning (DIL), and class-incremental learning (CIL) \citep{vandeVen2022}. The learning technique requiring the model to be updated incrementally on new classes in new tasks is called Class-Incremental Learning (CIL). The models trained under this setting should acquire the ability to differentiate all the previously observed classes. Although TIL also learns information from new classes in new tasks, it only requires the model to differentiate classes in the corresponding tasks. Thus, the models do not have the ability to discriminate classes within different tasks. On the other hand, DIL focuses on instances from different domains or distributions but with the same class labels. This research only concentrates on the scenario of CIL to address the issue of countless botanical origins of honey in the real world. 
	
	One challenge of CIL is catastrophic forgetting \citep{Kirkpatrick2016OvercomingCF}, i.e., directly training the network with new classes will interfere with the knowledge of former ones. This phenomenon can lead to overwriting old knowledge with new knowledge so that the model performance can decrease abruptly \citep{PARISI201954}. The other issue is loss of plasticity, i.e., the network trained on formerly known classes loses the ability to learn the new classes. The networks developed under the scenario of CIL must be plastic to integrate the new knowledge and stable to not negatively interfere with the formerly obtained information \citep{PARISI201954}. Hence, effectively addressing these issues and seeking trade-offs becomes the crucial challenge for developing CIL algorithms. A good model should balance capturing new classes' characteristics and retaining the formerly learned classes' concepts. This trade-off is known as the 'stability-plasticity dilemma' in neural systems \citep{Wang2024surveyCL}. 
	
	There have already been enormous research works for CIL tasks. They were categorized into different groups regarding various standards \citep{Masana2023classincremental, Wang2024surveyCL, Zhou2024CILsurvey, TIAN2024307}. Some of the widely applied categories include: (1) Data Replay \citep{Ratcliff1990ConnectionistMO}, which revisit former exemplars for current training tasks \citep{icarl2017Rebuffi, Chaudhry_2018_ECCV, Bang_2021_CVPR, liu2021rmm, Barry2023replay}; (2) Knowledge Distillation (KD) \citep{Hinton2015DistillingTK}, which allows the knowledge from former models to transfer to current ones, such as LwF \citep{LwF2016} and iCaRL \citep{icarl2017Rebuffi}; (3) Regularization-based methods, which add regularization terms to balance the former and current knowledge, such as GEM \citep{Lopez2017GEM}, UCIR \citep{ucir2019Hou}, PODNet \citep{PODNet2020}, COIL \citep{Zhou2021CoTransportFC}, EWC \citep{Kirkpatrick2016OvercomingCF}, SI \citep{Zenke2017ContinualLT}; and (4) Dynamic networks \citep{yoon2018lifelong}, which construct task-specific parameters or network architecture, such as AANets \citep{Liu2021AANets}, DER \citep{Der2021Yan}, Foster \citep{Foster2022Wang}, MEMO \citep{MEMO2023zhou}, DyTox \citep{DyTox2022Douillard}, and L2P \citep{L2P2022Wang}. 
	
	Furthermore, some works concentrate specifically on the issue of plasticity loss. For instance, \citep{Lyle2023understand} found that the changes in the curvature of the loss landscape is related to the loss of plasticity issue; Concatenated ReLUs (CReLUs) activation function \citep{pmlr-v232-abbas23a} was proposed to mitigate the networks' sparse activation footprint as more tasks conducted, which cause diminishing gradient issue; a simple L2 regularization toward initial parameters, which is named as L2 Init \citep{kumar2024l2init}, was proposed to mitigate plasticity loss by resetting neurons or parameter values; similarly, a variation of the backpropagation algorithm, i.e., continual backpropagation \citep{Dohare2024}, aims to inject plasticity into the network to maintain the variability by reinitializing less-used hidden neuron; Collaborative Continual Learning (CCL) \citep{Wang_2024_CVPR}, which is a collaborative learning based strategy, was presented to improve model plasticity; adaptive plasticity improvement (API) \citep{Liang_2023_CVPR} was introduced to evaluate a network's plasticity and adaptively improve the its plasticity for learning a new task if necessary. 
	
	There are applications of CIL techniques in real-world tasks, e.g., plant disease classification \citep{LI2024109211}, and human activity recognition \citep{WANG2025126893}. Some works utilizing CIL methods on HSI datasets. For example, IL based on one-class learning (OCL) was applied to identify maize seed varieties \citep{ZHANG2022107153}; LwF was used for crop growth parameters estimation and  leaf nitrogen diagnosis \citep{DU2023108356}; A few-shot CIL was implemented using the Replay training strategy to categorize Chrysanthemum \citep{CAI2023108371}; Knowledge distillation and replay were utilized to recognize plant species \citep{Chien2024}; A linear programming incremental learning classifier was proposed to classify HSI remote sensing data \citep{Bai2022linear}; Previous label replay strategy (PLRS) was adopted to classify remote sensing HSI data by synthesizing fused labels from previous model with pseudo-labels of the current target \citep{Dong2024plrs}. However, most of the works in CIL field only experimented on public dataset such as Cifar, ImageNet, Indian Pine, Pavia University, and other public remote sensing datasets. The honey botanical origin classification task has never been studied with CIL method. 
	
	This work presents implementing different CIL techniques on the honey hyperspectral imaging dataset to learn to discriminate honey botanical origins incrementally. Moreover, we also developed a method to utilize a variation of the backpropagation algorithm, ie, continual backpropagation (CB) \citep{Dohare2024}, which was designed to inject plasticity into the network to maintain variability. CB is combined with other CIL techniques to address the issue of loss of plasticity by reinitializing a small portion of less-used hidden units during incremental training \citep{Dohare2024}. To the best of our knowledge, this is the first work that examines the performance of CIL algorithms on real-world honey HSI data. In addition, this is also the first work that combines CB with other CIL techniques and examines their effects on real-world HSI honey datasets. The main contribution of this study would be:
	
	\begin{enumerate}
		\item We performed various CIL algorithms on real-world honey HSI dataset.
		
		\item We proposed to integrate continual backpropagation (CB) with different CIL methods and conducted sensitivity analysis for different hyperparameter combinations of CB. 
		
		\item We conducted experiments on honey HSI dataset to compare the performance between vanilla CIL methods and CIL + CB methods.
		
		\item We also analyzed the kernel weight distributions of the networks used for different CIL methods to display the their effects on the neural network.
	\end{enumerate}
	
	The CIL algorithms utilized in this research are introduced in Section \ref{cilmethod}, and CB is described in Section \ref{continualbackprop}. The experiment results are presented in Section \ref{cilresult}. The classification results of different CIL methods are discussed in Section \ref{resultcil}. Section \ref{cbhyper} performs the sensitivity analysis of different combinations of CB's hyperparameters. The classification results of CIL methods and CIL + CB methods are compared in Section \ref{cilcbcomp}. The kernel weight distributions of the last layer for both CIL and CIL + CB techniques are analyzed in Section \ref{cilkwdanalysis}.
	
	\section{Materials and Methods}
	This section briefly introduces the collection of honey samples by \citep{NOVIYANTO2019129} and acquisition of honey samples' hyperspectral imaging. A dataset is developed utilizing the hyperspectral images acquired by \citep{noviyanto2017honey} with different data sampling and cleaning methods, as introduced in \citep{Zhang2023optimize}. Section \ref{cilmethod} briefly introduces the different class-incremental learning (CIL) methods implemented for our experiments. And continual backpropagation (CB) is discussed in Section \ref{continualbackprop}.
	
	\subsection{Datasets}
	
	The honey products were from 25 botanical origins, which were produced by 11 different brands in New Zealand \citep{NOVIYANTO2020109684}. This dataset contains multiple mono-floral honey, such as Blue Borage, Borage Field, Clover, Honeydew, Pohutukawa, Rata, Rewarewa, Tawari, and various Mānuka honey. The Mānuka honey, which is graded by the Unique Mānuka Factor (UMF) system, includes UMF5+, 10+, 12+, 13+, 15+, 18+, 20 + and 22+ types, and also one ungraded Mānuka honey. Besides the monofloral honey types, the samples included ungraded Mānuka, Mānuka Blend, Wildland, Wildflower, and Multifloral honey types \citep{noviyanto2017honey}. An oven heated the honey samples to dissolve crystals at 40$\degree$C in closed containers overnight. The liquified honey samples were split into six acquisitions, each containing 7 grams of honey. Each acquisition's 7 grams of honey produced more or less 5 mm thickness in a 35-mm-diameter lime glass container. The thickness of each honey acquisition should be standard to minimize the spectral information variations caused by sample thickness \citep{noviyanto2017honey}. According to \citep{noviyanto2017honey}, the standard deviation of thickness is 0.0063 cm, which is too low to affect the spectral information significantly. 
	
\begin{figure}[ht]
		\centering
		\includegraphics[scale=1.0]{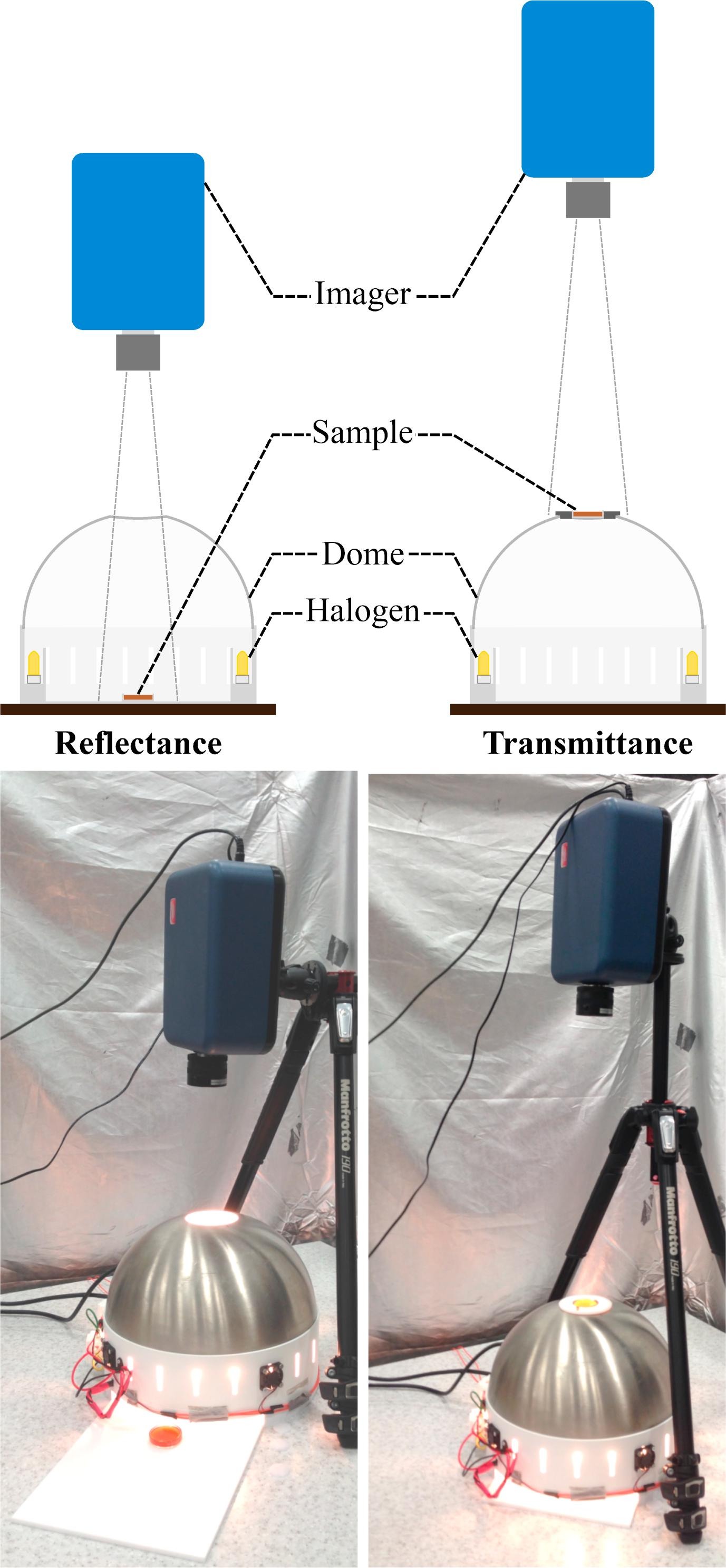}
		\caption{\footnotesize Reflectance and transmittance sensing modes using the ring light configuration with a dome. \citep{NOVIYANTO2019129}}
		\label{fig:spect}
	\end{figure}

	The hyperspectral imaging device used in this investigation was a SOC710VP Hyperspectral Camera from Surface Optic Corporation in California, USA, with a Schneider-Kreuznach Xenoplan 35mm lens. Each honey sample was stored in a lime green glass jar with an array of halogen bulbs arranged in a circular shape surrounded by a dome, as illustrated in Figure \ref{fig:spect}. Pushbroom line-scanning was used to acquire the hyperspectral image data cubes \citep{NOVIYANTO2019129}. Each hyperspectral image contains 128 spectral bands from 399.40 to 1063.79 nm (i.e., Visible to Near-Infrared spectrum, VIS-NIR) with approximately 4.9 nm spectral resolution and 520$\times$696 pixels spatial resolution with 12-bit data per pixel (0-4095 intensity value). There are 25 samples for each acquisition, and the total number of data samples in the hyperspectral imaging dataset is 8675. HSI data pixels were randomly selected, then the noisy points, such as bubbles, were removed with clustering method. The sampling and cleaning details were introduced in \cite{Zhang2023optimize}. All the selected samples were normalized using SNV method \citep{snv1989}. Because this work is designed for class-incremental learning tasks, the honey's botanical origins should be increased according to the number of tasks. Therefore, the botanical origin classes used for each incremental learning task are displayed in Table \ref{cildata}. 
	
	\begin{table}
		\hspace{-0.3in}
		{\small 
			\begin{tabular}{||c c c c c||}
				\hline
				\multicolumn{5}{||c||}{\textbf{Honey botanical origins and Sample Numbers for Each Class Incremental Learning Task}} \\
				\hline 
				\textbf{\textit{0}} & \textbf{\textit{1}} & \textbf{\textit{2}} & \textbf{\textit{3}} & \textbf{\textit{4}} \\
				\hline
				0 BBLiquid: 300 & 5 Field+Tawari: 150 & 10 MānukaUMF10: 900 & 15 MānukaUMF20: 275 & 20 Rata: 300 \\
				\hline
				1 BorageField: 150 & 6 Honeydew: 150 & 11 MānukaUMF12: 150 & 16 MānukaUMF22: 150 & 21 Rewarewa: 900 \\
				\hline
				2 Clover: 600 & 7 Kamahi: 300 & 12 MānukaUMF13: 150 & 17 MānukaUMF5: 900 & 22 Tawari: 450 \\
				\hline
				3 Clover Cream: 150 & 8 Mānuka: 450 & 13 MānukaUMF15: 600 & 18 Multifloral: 300 & 23 Wildflower: 150 \\
				\hline
				4 Clover Liquid: 150 & 9 MānukaBlend: 450 & 14 MānukaUMF18: 150 & 19 Pohu: 300 & 24 Wildland: 150 \\
				\hline
			\end{tabular}
		}
		\caption{Honey Sample Number for CIL Tasks}
		\label{cildata}
	\end{table}
	
	\subsection{Class-Incremental Learning Methods}\label{cilmethod}
	
	Among the numerous research works targeting the CIL area, we selected nine representative methods to experiment with the honey HSI dataset, including Finetune as the baseline, LwF \citep{LwF2016}, EWC \citep{Kirkpatrick2016OvercomingCF}, Replay \citep{Ratcliff1990ConnectionistMO}, iCaRL \citep{icarl2017Rebuffi}, WA \citep{WA2020Zhao}, DER \citep{Der2021Yan}, Foster \citep{Foster2022Wang}, and MEMO \citep{MEMO2023zhou}. Then, these algorithms were combined with continual backpropagation algorithm \citep{Dohare2024} to examine the effects on classification performance. These CIL methods can be grouped into categories concerning various specific focuses, which are not mutually exclusive. They were categorized according to a survey on CIL \citep{Zhou2024CILsurvey}, whose codes of these CIL techniques were also modified to develop our work. For all these CIL algorithms, the dimension of the final fully connected layer was extended to include the number of classes seen so far, as shown in Table \ref{cnnarchinorm}. 
	
	\begin{enumerate}
		\item The \textbf{Finetune} method, which was set as the baseline, directly used the network trained with the samples from new classes to predict the samples from all the classes seen so far. The Finetune method, which was set as the baseline, directly used the network trained with the samples from new classes to predict the samples from all the classes seen so far. Given the current dataset $\mathcal {D}^{b}$, the loss function of Finetune can be denoted as:
		
		\begin{equation}\label{finetune} 
			\textstyle \mathcal {L}= \sum _{(\mathbf{x},y)\in \mathcal {D}^{b}} \ell (f(\mathbf{x}),y) 
		\end{equation}
		
		where $\ell (\cdot,\cdot )$ is the cross-entropy loss, $\mathbf{x}$ is the input data, and $y$ is the corresponding class label. Finetune is used as the baseline method because there is no modification to the network architecture or loss function and no exemplars in the training set. It only used training samples from the current task to learn new knowledge for the current task. 
		
		\item One type of CIL technique is Parameter Regularization, based on the assumption that the contributions of network parameters are not equal. Thus, Parameter Regularization methods attempted to evaluate the importance of different parameters to the network and to maintain former knowledge by keeping the important ones static \citep{Zhou2024CILsurvey}. \textbf{EWC} \citep{Kirkpatrick2016OvercomingCF} addressed the parameter regularization by maintaining an importance matrix $\Omega$. Given the $k$-th model parameters $\theta_k$, the importance of $\theta_k$ is represented by $\Omega _{k}\geq 0$, where the larger $\Omega _{k}\geq 0$ indicates more important $\theta_k$. EWC adds a regularization term to loss function in Eq. (\ref{finetune}):
		
		\begin{equation}\label{ewc} 
			\mathcal {L}= \ell (f(\mathbf{x}),y)+ \frac{1}{2}\lambda \sum _{k}\Omega _{k}(\theta _{k}^{b-1}-\theta _{k})^{2} 
		\end{equation}
		
		where $\theta _{k}^{b-1}$ is the $k$-th parameter after learning the last task $\mathcal {D}^{b-1}$, Thus, the parameter drift from the last task can be acquired by $(\theta _{k}^{b-1}-\theta _{k})^{2}$, and $\Omega _{k}$ keeping important parameters from shifting from the last stage. The importance matrix $\Omega$ was estimated by Fisher information matrix. Other methods that were designed to improve this optimization dynamics, such as SI \citep{Zenke2017ContinualLT}, which proposed to estimate $\Omega$ in an online manner, is beyond the scope of this work. 
		
		\item Knowledge Distillation (KD) \citep{Hinton2015DistillingTK}, which enables the knowledge transfer from a teacher model to the student model, can also be deployed for CIL tasks by keeping the old network $f^{b-1}$ from the previous learning task and the new model $f$ updated for the current task. Similar to Eq (\ref{ewc}), \textbf{LwF} \citep{LwF2016} utilized knowledge distillation in CIL by adding the regularization term:
		
		\begin{equation}\label{lwf} 
			\mathcal {L}= \underbrace{ \vphantom{\sum _{k=1}^{|\mathcal {Y}_{b-1}|}} \ell (f(\mathbf{x}),y)}_{\text{New Classes}}+ \lambda\underbrace{ \sum _{k=1}^{|\mathcal {Y}_{b-1}|}- \mathcal {S}_{k}({f}^{b-1}(\mathbf{x})) \log \mathcal {S}_{k}(f(\mathbf{x}))}_{{\text{Remembering Old Classes}}} 
		\end{equation}
		
		where $\mathcal {S}_{k}$ is the Softmax activation function, $\lambda$ is used to control the contribution of knowledge distillation loss from old network $f^{b-1}$, which is frozen while training new task. Eq (\ref{lwf}) balances a trade-off between the old network, which maintains old knowledge, and new classes, which aims to learn new concepts. iCaRL \citep{icarl2017Rebuffi} extends LwF by utilizing the exemplars, which the new network can use to recreate former knowledge during incremental training.

		\item Another type of incremental learning technique is applying an exemplar set, i.e., Data Replay. \textbf{Replay} is performed by saving a small fraction of samples from previously known classes for network training \citep{Ratcliff1990ConnectionistMO}. \textbf{iCaRL} \citep{icarl2017Rebuffi} extends LwF by utilizing an exemplar selection strategy called herding, which aims to select the most representative instances from previously seen classes. Herding calculates the center of each class and selects the samples that are more closer to the class center within the memory limit. However, despite the numerous methods to construct the exemplar set for CIL training \citep{Zhou2024CILsurvey, MEMO2023zhou, liu2021rmm}, we did not apply any specific exemplar set selection method. Instead, we randomly selected 20 exemplars from each known class and concatenated them to the training set of the current CIL task. The efficiency and effectiveness of exemplar selection techniques are not concerns for this work. 
		
		\item There were also works attempting to explore the abnormal behaviors of networks during CIL training. UCIR \citep{ucir2019Hou} finds that the incrementally trained networks tend to predict samples as newly encountered classes because the weight norm of new classes is significantly larger than that of old ones. Thus, \textbf{WA} \citep{WA2020Zhao} normalizes the weight corresponding to the old tasks after each incremental learning task:
		
		\begin{equation}\label{wa}
			\hat{W}_{old} = W_{old} * \frac{\frac{1}{m}||W_{old}||}{\frac{1}{n}||W_{new}||}
		\end{equation}
		
		where $W_{old}$ represents the weight of old tasks, $\frac{1}{m}||W_{old}||$ denotes the mean weight norm for old tasks, and $\frac{1}{n}||W_{new}||$ is the mean weight norm for new tasks.

		\item Another group of CIL techniques is Dynamic Networks \citep{yoon2018lifelong}, which are designed to dynamically expand the network's representation ability for incremental learning tasks. \textbf{DER} \citep{Der2021Yan} was proposed to expand a new backbone for each new task and aggregate the features with a larger, fully connected layer. The old backbone is frozen to maintain former knowledge for new learning tasks.
		
		\begin{equation}\label{der}
			\mathcal {L}= \sum _{k=1}^{|\mathcal {Y}_{b}|}- {I}(y=k) \log \mathcal {S}_{k}(W_{new}^\top [\bar{\phi }_{old}(\mathbf{x}),\phi _{new}(\mathbf{x})]) 			
		\end{equation}
		
		where $\bar{\phi }_{old}(\mathbf{x})$ denotes the frozen old backbone trained for former tasks, $\phi _{new}$ is the new backbone for the current task, and $W_{new}$ is the new FC layer on the aggregated features, which is denoted as $[ \cdot,\cdot ]$.

		\item One major drawback of DER is the infinitely growing memory size for saving networks. \textbf{FOSTER} \citep{Foster2022Wang} suggests that features learned by former incremental learning tasks can be compressed by knowledge distillation. The number of backbones is limited to the teacher network and student network:
		
		\begin{equation}\label{foster} 
			\min _{f_{s}(\mathbf{x})} \operatorname{KL} \left( \mathcal {S}\left( f_{t}(\mathbf{x}) \right) \Vert \mathcal {S}\left( f_{s}(\mathbf{x}) \right) \right)
		\end{equation}
		
		where $f_{t}(\mathbf{x})=W_{new}^\top [\phi _{old}(\mathbf{x}),\phi _{new}(\mathbf{x})]$ is the frozen teacher network, and $f_{s}(\mathbf{x})=W^\top \phi (\mathbf{x})$ is the newly initialized student model. The number of backbones is limited to the teacher network and student network. Eq (\ref{foster}) aims to find the student model $f_{s}$ with the same discrimination ability as the teacher model $f_{t}$ by minimizing the discrepancy between them.   
		
		\item \textbf{MEMO} \citep{MEMO2023zhou} finds that shallow layers of networks from different CIL learning stages are more similar than deep layers, indicating that deep layers are specific to the different tasks. Hence, MEMO proposes to decouple the backbone: $\phi (\mathbf{x})=\phi _{s}(\phi _{g}(\mathbf{x}))$, where specialized block $\phi _{s}$ corresponds to the deep layers in the network, and generalized block $\phi _{g}$ corresponds to the shallow layers. MEMO only expands specialized blocks $\phi _{s}$ and transforms Eq (\ref{der}) into:
		
		\begin{equation}
			\sum _{k=1}^{|\mathcal {Y}_{b}|}- {I}(y=k) \log \mathcal {S}_{k}(W_{new}^\top [{\phi _{s}}_{old}(\phi _{g}(\mathbf{x})),{\phi _{s}}_{new}(\phi _{g}(\mathbf{x}))]) 
		\end{equation}
		
		which indicates that deep layers can be designed for each CIL task based on the shared shallow layers $\phi _{g}(\mathbf{x})$. We used one fully connected layer and the last convolutional layer as the task-specific deep layers. All the other convolutional layers were considered as the general feature extraction layers.

	\end{enumerate}

	\subsection{Continual backpropagation}\label{continualbackprop}
	
\begin{figure}[ht]
		\hspace{-0.6in}
		\begin{subfigure}[c]{0.45\textwidth}
			\hspace{-0.35in}
			\includegraphics[scale=0.38]{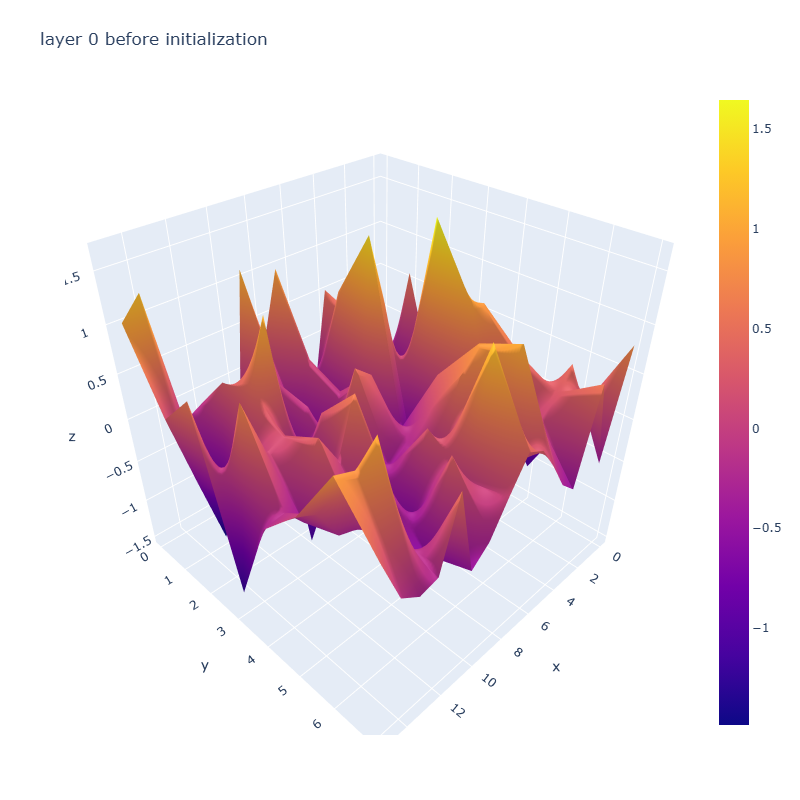}
			\caption{}
			\label{cbpre}
		\end{subfigure}
		\hfill
		\begin{subfigure}[c]{0.45\textwidth}
			\hspace{-0.55in}
			\includegraphics[scale=0.38]{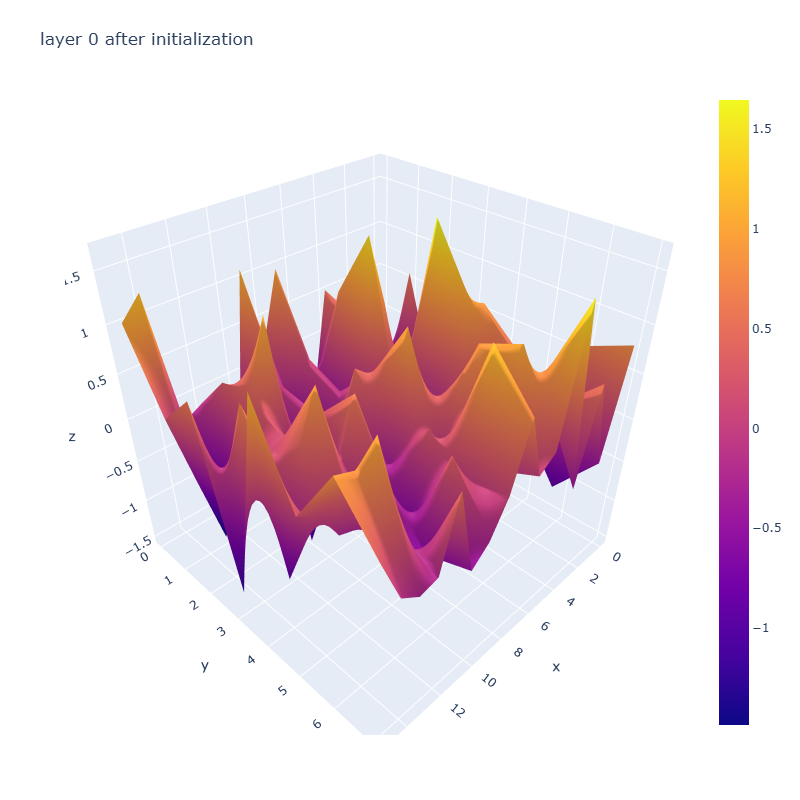}
			\caption{}
			\label{cbafter}
		\end{subfigure}
	    \label{cb3d}
	    \caption{(a).3D Plot of a Convolutional Layer of Size (32,1,4) before Reinitialization; (b).3D Plot of a Convolutional Layer of Size (32,1,4) after Reinitialization}
	\end{figure}

	Continual backpropagation \citep{Dohare2024} randomly reinitializes a proportion of low-utility units in the network. The utility measure is defined to measure the magnitude and contribution of the product of units’ activation and outgoing weight. If the contribution of a hidden unit is small, this hidden unit is less useful than other hidden units. The contribution utility ${{\bf{u}}}_{l}[i]$ of a $i$th hidden unit in layer $l$ at time $t$ is designed as the sum of the utilities, which is measured as a running average of contributions with a decay rate $\eta$. 
	
	\begin{equation}
		{{\bf{u}}}_{l}[i]=\eta \times {{\bf{u}}}_{l}[i]+(1-\eta )\times | {{\bf{h}}}_{l,i,t}| \times \mathop{\sum }\limits_{k=1}^{{n}_{l+1}}| {{\bf{w}}}_{l,i,k,t}| ,
	\end{equation}
	
	where ${{\bf{h}}}_{l,i,t}$ is the output of the $i$th hidden unit in layer $l$ at time $t$, ${{\bf{w}}}_{l,i,k,t}$ is the weight connecting the $i$th unit in layer $l$ to the $k$th unit in layer $l+1$ at time $t$, and $n_{l+1}$ is the number of units in layer $l+1$. The outgoing weights of a reinitialized hidden unit are set to zero, ensuring that the new hidden units do not affect the already learned function. In order to prevent the newly initialized hidden units, which have zero utility, from immediate reinitialization, they are protected by setting a maturity threshold $m$ number of updates. We call a unit mature if its age is more than $m$. The age of a hidden unit is related to the number of epochs, the number of batches for each learning step, and the number of hidden layers. If the maturity threshold $m$ is set to 2000, given a network with 24 hidden layers and five mini-batches for the training dataset, the hidden units can be mature every $16.67 = 2000 / (24 * 5)$ epochs. A fraction, which is determined by the replacement rate of $\rho$, of mature units is reinitialized in every layer. In order to evaluate the effects of different hyperparameters on CB, we performed a sensitivity analysis for the replacement rate of $\rho$ and maturity threshold $m$, as presented in Section \ref{cbhyper}. For each CIL method with CB, the best combination of replacement rate $\rho$ and maturity threshold $m$ were shown in Table \ref{cbbest}. According to the definition of the different dynamic neural networks, DER applied CB on the newly added network for current task, FOSTER utilized CB on the student network, and MEMO implemented CB on the generalized block. Other CIl methods implemented CB on all the convolutional layers, linear layers, and batchnorm layers except the last fully connected layer, which was updated concerning each CIL task. Figs \ref{cbpre} and \ref{cbafter} show the 3D plots of a convolutional layer with shape (32,1,4) before and after the neuron reinitialization with replacement rate of 0.1. 
	
	\subsection{Performance Evaluation}\label{pe}
	The weighted average F1 score was applied to evaluate the performance of the classification models. In order to ensure that the scores of the exemplar set do not significantly affect the classification outcomes, the weighted average F1-score computes the weighted mean of the F1 scores based on the number of samples of each class. The F1 score, precision, and recall scores were calculated as:
	
	\begin{equation}\label{f1}
		\begin{split}
			& \text{F1} = \frac{2 \times (\text{precision} \times \text{recall})}{\text{precision + recall}} \\
			& \text{precision} = \frac{\text{TP}}{\text{TP + FP}} \\
			& \text{recall} = \frac{\text{TP}}{\text{TP + FN}} \\
		\end{split}
	\end{equation}
	
	\noindent True positives (TP), false positives (FP), and false negatives (FN) were used to compute these scores. Recall assesses the classifier's capacity to identify every positive sample, whereas precision gauges its capacity to avoid labeling a negative sample as positive.

	\section{Results and Discussion}\label{cilresult}
	
	\subsection{Experiment Setting and Backbone model}\label{backbone}
	All the experiments were run on a laptop with NVIDIA Geforce GTX3060 6.0 GB GPU and 16.0GB RAM. Five rounds of experiments were conducted for each sub-task to obtain the average and standard deviation of classification scores. For each class-incremental learning task, the honey dataset used 70\% for training, 15\% for validation, and 15\% for testing. The test results were saved based on the best validation accuracy. We started with training on five botanical origin classes and then added more, five at a time, until all 25 were available. The botanical origin classes used for each incremental learning task are displayed in Table \ref{cildata}. Unlike other methods for constructing the exemplar set for CIL training, such as herding and reinforced memory management \citep{Zhou2024CILsurvey, MEMO2023zhou, liu2021rmm}, we randomly selected 20 exemplars from each known class without exemplar memory size limit. All the selected exemplars were concatenated with the training set of the current CIL task. After each addition, the networks were trained, and performance was measured for all available classes. 
	
\begin{figure}[ht]
		\begin{center}
			\includegraphics[width=1.0\textwidth]{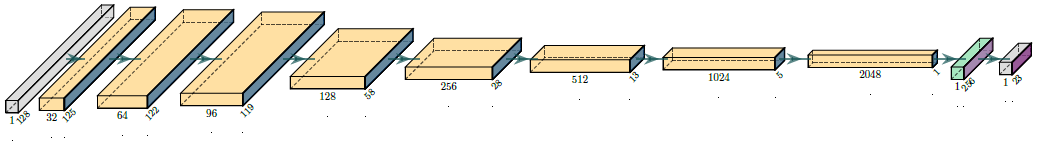}
		\end{center}
		\caption{CNN Architecture as the Base Model.}
		\label{cnnarchi}
	\end{figure}

	For all the experiments, a 1D-CNN is used as the backbone for all the CIL tasks, as shown in Fig. \ref{cnnarchi}. The input of the network is hyperspectral image pixels with 128 spectral bands. The parameters of kernel size / stride step / padding, output channel number, and output shape of each layer are shown in Table \ref{cnnarchinorm}. Adam optimizer is applied with 0.0001 learning rate. Each method was trained for 50 epochs on each incremental learning task with a batch size of 256. Other hyperparameters, including the $\lambda$ to balance the trade-off between old and new knowledge in LwF and iCarl, utilized the default setting of experiments in \citep{Zhou2024CILsurvey}.
	
	\begin{table}[ht]
		\centering
		\begin{tabular}{ |p{1.5cm}|c|c| }
			\hline
			\multicolumn{3}{|c|}{CNN Model Architecture} \\
			\hline
			& parameter & output shape \\
			\hline
			Input & - & 1x128  \\
			\hline
			Layer 0 & 4/1/0, 32 & 32x125  \\
			\hline
			Layer 1 & 4/1/0, 64 & 64x122  \\
			\hline
			Layer 2 & 4/1/0, 96 & 96x119 \\
			\hline
			Layer 3 & 4/2/0, 128 & 128x58  \\
			\hline
			Layer 4 & 4/2/0, 256 & 256x28  \\
			\hline
			Layer 5 & 4/2/0, 512 & 512x13 \\
			\hline
			Layer 6 & 4/2/0, 1024 & 1024x5  \\
			\hline
			Layer 7 & 4/2/0, 2048 & 2048x1  \\
			\hline
			\multicolumn{3}{|c|}{Fully Connected  256} \\
			\hline
			\multicolumn{3}{|c|}{Fully Connected  (5, 10, 15, 20, 25)} \\
			\hline
			
		\end{tabular}
		\caption{CNN Model Architectures, the parameter column shows kernel size/stride step/padding, and output channels for each convolutional layer respectively. The output neurons of the final fully connected layer increase according to different CIL tasks.}
		\label{cnnarchinorm}
	\end{table}
	
	\subsection{Classification results of different CIL methods}\label{resultcil}
	
	The classification results shown in Tables \ref{tab1} and \ref{tab2} are the mean values and standard deviations of test-weighted average F1 scores from the five rounds of experiments. The model sizes and each epoch's computational time of different CIL algorithms are presented in Table \ref{tabsize}. 
	
	\subsubsection{Model sizes and computational time}
	The model size and computational time for each CIL technique were presented in Table \ref{tabsize}. The training time differences between vanilla CIL methods and CIL + CB methods are not significant. Finetune had the lowest computational time for each CIL task because there was no extra computation for this method. Compared to other non-dynamic network CIL techniques, EWC has a higher computational cost because it has to compute a Fisher information matrix as the parameter regularization. Replay had a relatively higher computational cost than Fintune because Replay appended exemplars to the training set. The model sizes and each epoch's training time of dynamic network algorithms, such as DER and MEMO, significantly increase as more tasks are conducted. DER appends an additional neural network to the previous networks and only trains the new network. The model sizes of DER increase linearly with the number of CIL tasks. The model sizes of MEMO increase as the task-specific blocks extend with more CIL tasks, while the general block remains the same regardless of CIL tasks. 
	
	\begin{table}[ht]
		\caption{Incremental Learning Model Size and Computational Cost. In each cell, the model size, each epochs's training time for vanilla CIL method and CB method are shown.}
		\hspace{-0.3in}
		\begin{tabular}{|c|p{3cm}|p{3cm}|p{3cm}|p{3cm}|p{3cm}|}
			\hline
			\textbf{CIL}&\multicolumn{5}{c|}{\textbf{Model Sizes and Each Epoch's Training Time of Incremental Task No.}} \\
			\cline{2-6} 
			\textbf{Method} & \textbf{\textit{0}} & \textbf{\textit{1}} & \textbf{\textit{2}} & \textbf{\textit{3}} & \textbf{\textit{4}} \\
			\hline
			Finetune & 46MB, 0.11s, cb:0.11s & 46MB, 0.13s, cb:0.13s & 46MB, 0.15s, cb:0.15s & 46MB, 0.15s, cb:0.15s & 46MB, 0.16s, cb:0.16s \\
			\cline{2-6}
			LwF & 46MB, 0.11s, cb:0.11s  & 46MB, 0.16s, cb:0.16s & 46MB, 0.19s, cb:0.19s & 46MB, 0.19s, cb:0.19s  & 46MB, 0.20s, cb:0.20s \\
			\cline{2-6}
			EWC & 46MB, 0.23s, cb:0.23s  & 46MB, 0.37s, cb:0.37s & 46MB, 0.38s, cb:0.38s & 46MB, 0.42s, cb:0.42s  & 46MB, 0.43s, cb:0.43s \\
			\cline{2-6}
			Replay & 46MB, 0.10s, cb:0.10s & 46MB, 0.13s, cb:0.13s & 46MB, 0.18s, cb:0.18s  & 46MB, 0.17s, cb:0.17s  & 46MB, 0.18s, cb:0.18s \\
			\cline{2-6}
			iCarl & 46MB, 0.10s, cb:0.10s & 46MB, 0.17s, cb:0.17s & 46MB, 0.22s, cb:0.22s  & 46MB, 0.23s, cb:0.23s & 46MB, 0.23s, cb:0.23s \\
			\cline{2-6}
			WA & 46MB, 0.11s, cb:0.11s & 46MB, 0.16s, cb:0.16s  & 46MB, 0.22s, cb:0.22s  & 46MB, 0.23s, cb:0.23s  & 46MB, 0.23s, cb:0.23s \\
			\cline{2-6}
			DER & 46MB, 0.11s, cb:0.11s & 92MB, 0.19s, cb:0.19s & 138MB, 0.29s, cb:0.29s  & 184MB, 0.34s, cb:0.34s  & 230MB, 0.38s, cb:0.38s \\
			\cline{2-6}
			FOSTER & 46MB, 0.11s, cb:0.11s  & 92MB, 0.20s, cb:0.20s & 92MB, 0.27s, cb:0.27s & 92MB, 0.28s, cb:0.28s  & 92MB, 0.30s, cb:0.30s \\
			\cline{2-6}
			MEMO & 46MB, 0.11s, cb:0.11s  & 80MB, 0.19s, cb:0.19s  & 115MB, 0.31s, cb:0.31s  & 150MB, 0.37s, cb:0.37s & 185MB, 0.43s, cb:0.43s \\
			\hline
		\end{tabular}
		\label{tabsize}	
		
	\end{table}
	
	\subsubsection{Performance of CIL methods}
	The performance of different CIL algorithms can be compared in Table \ref{tab1}. The Retrain in the first row presents the results for fully retraining the network on all the seen classes. Clearly, the methods using exemplars, including Replay, iCarl, WA, DER, FOSTER, and MEMO, obtain better F1 scores than Finetune, LwF, and EWC. Moreover, different methods display various performances on different CIL tasks. For instance, LwF scored higher on task 1 than EWC and Finetune, while it obtained a lower F1 score than EWC on task 2. EWC had lower F1 scores than the baseline Finetune method on tasks 3 and 4. A similar pattern can also be observed for methods with exemplars. Although DER acquired the highest scores on tasks 2 and 4, FOSTER performed better on tasks 1 and 3.  
	
	However, of all the six algorithms using exemplars, some obtained lower classification results compared to the Replay method. In Table \ref{tab1}, iCarl, WA, and MEMO had lower F1 scores than Replay on tasks 1, 2, and 3, while they only achieved higher scores on task 4. In addition, FOSTER also had a lower F1 score than Replay on task 2, but it achieved much higher scores on tasks 1, 3, and 4. 
	
\begin{figure}
		\centering
		\includegraphics[scale=0.6]{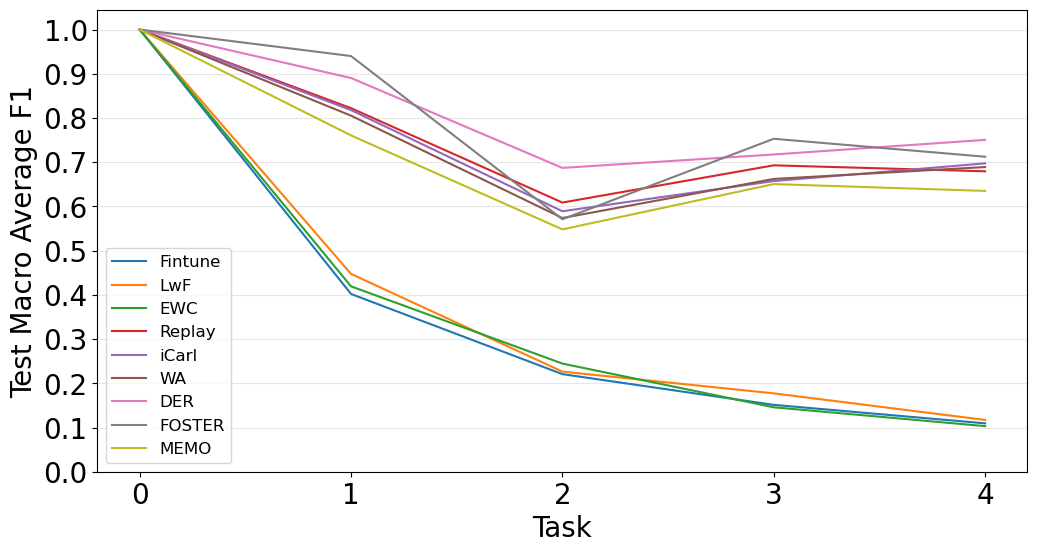}
		\caption{Comparison of different class-incremental learning algorithms. The mean values for the test weighted average F1 scores of the five rounds of experiments were shown for each class-incremental learning task. }
		\label{wf1}
	\end{figure}

	In Fig. \ref{wf1}, the algorithms using exemplars showed a distinct pattern when comparing performance among different tasks: unlike the three methods without exemplars, which experienced performance decline as the number of botanical origin classes increased, the algorithms using exemplars presented performance improvement from task 2 to 3. This phenomenon might be caused by the similarity of honey botanical origins between tasks 2 and 3, which mainly contain graded Mānuka honey in both tasks, as shown in Table \ref{cildata}.

	\begin{table}[ht]
		
		\begin{center}
			\begin{tabular}{|c|c|c|c|c|c|}
				\hline
				\textbf{CIL}&\multicolumn{5}{c|}{\textbf{Weighted Average F1 Score of Incremental Task No.}} \\
				\cline{2-6} 
				\textbf{Method} & \textbf{\textit{0}} & \textbf{\textit{1}} & \textbf{\textit{2}} & \textbf{\textit{3}} & \textbf{\textit{4}} \\
				\hline
				Retrain & 100.00$\pm$0.00\% & 100.00$\pm$0.00\% & 100.00$\pm$0.00\% & 99.78$\pm$0.08\% & 99.83$\pm$0.19\% \\
				\cline{2-6}
				Finetune & 100.00$\pm$0.00\% & 40.20$\pm$7.26\% & 22.07$\pm$3.44\% & 15.12$\pm$2.00\% & 10.94$\pm$1.14\% \\
				\cline{2-6}
				LwF & 100.00$\pm$0.00\% & 44.74$\pm$8.84\% & 22.64$\pm$5.70\% & 17.72$\pm$3.57\% & 11.71$\pm$1.35\% \\
				\cline{2-6}
				EWC & 100.00$\pm$0.00\% & 41.92$\pm$4.16\% & 24.46$\pm$2.80\% & 14.57$\pm$2.07\% & 10.31$\pm$1.16\% \\
				\cline{2-6}
				Replay & 100.00$\pm$0.00\% & 82.22$\pm$5.26\% & 60.87$\pm$6.71\% & 69.29$\pm$4.06\% & 67.94$\pm$4.98\% \\
				\cline{2-6}
				iCarl & 100.00$\pm$0.00\% & 81.78\%2.15\% & 58.91$\pm$7.20\% & 65.73$\pm$4.14\% & 69.74$\pm$4.33\% \\
				\cline{2-6}
				WA & 100.00$\pm$0.00\% & 80.52$\pm$4.05\% & 57.38$\pm$8.38\% & 66.21$\pm$6.13\% & 68.91$\pm$6.56\% \\
				\cline{2-6}
				DER & 100.00$\pm$0.00\% & 89.04$\pm$4.88\% & 68.71$\pm$4.61\% & 71.75$\pm$2.55\% & 75.05$\pm$4.28\% \\
				\cline{2-6}
				FOSTER & 100.00$\pm$0.00\% & 94.01$\pm$2.63\% & 57.09$\pm$7.49\% & 75.30$\pm$4.69\% & 71.24$\pm$2.45\% \\
				\cline{2-6}
				MEMO & 100.00$\pm$0.00\% & 76.06$\pm$5.32\% & 54.81$\pm$7.14\% & 65.05$\pm$7.20\% & 63.51$\pm$7.37\% \\
				\hline
			\end{tabular}
			\caption{Classification Results for different CIL algorithms. Retrain represents retraining on all the classes seen so far.}
			\label{tab1}
		\end{center}
	\end{table}
	
	\subsubsection{Confusion Matrix}\label{cilcf}
	
	Figs. \ref{der44}, \ref{foster44}, \ref{memo44}, \ref{replay44}, \ref{icarl44}, and \ref{wa44} display the confusion matrix of DER, FOSTER, MEMO, Replay, iCarl, and WA's classification results on task 4 from a randomly chosen experiment round. In Fig. \ref{der44}, it is clear that DER tends to misclassify Mānuka, MānukaUMF5, and MānukaUMF10 as Rewarewa honey. In addition, DER also confused some MānukaUMF10 samples with MānukaUMF5 and Clover honey samples with Tawari. In Fig. \ref{foster44}, FOSTER also misclassifies some MānukaUMF10 samples as MānukaUMF5 and a few MānukaUMF15 / MānukaUMF18 samples as other UMF graded Mānuka honey types. There are also honey samples of Mānuka, MānukaUMF10, and MānukaUMF12 are confused with Rewarewa honey. Some Clover honey samples are falsely identified as Pohu and Tawari honey as well. Fig. \ref{memo44} shows that MEMO confuses Mānuka, MānukaBlend, MānukaUMF5, MānukaUMF10, and MānukaUMF12 with Rewarewa honey, and Clover and Kamahi with Tawari honey. A similar pattern can also be observed in Replay, iCarl, and WA methods, as shown in Figs \ref{replay44}, \ref{icarl44}, and \ref{wa44}. Generally speaking, we can tell from figures that the incrementally trained models are inclined to misclassify Mānuka and UMF graded Mānuka honey as other UMF graded Mānuka honey and Rewarewa honey, while Clover honey can also be confused with Tawari honey.

	\subsection{Hyperparameter Sensitive Analysis of CIL + CB}\label{cbhyper}
	
	In order to analyse the effects of different hyperparameters on CB, a sensitivity analysis was conducted on two essential hyperparameters, i.e. replacement rate and maturity threshold. The weighted average F1 scores on different CIL tasks are shown in Tables \ref{hpfinetune} for Finetune  + CB, \ref{hplwf} for LwF + CB, \ref{hpewc} for EWC + CB, \ref{hpreplay} for Replay + CB, \ref{hpicarl} for iCarl + CB, \ref{hpwa} for WA + CB, \ref{hpder} for DER + CB, \ref{hpfoster} for FOSTER + CB, and \ref{hpmemo} for MEMO + CB. 
	
	Table \ref{hpfinetune} shows the parameter tunning results of Finetune + CB. In task 1, the replacement rate/maturity threshold combination of 0.001 / 500 achieved the highest weighted average F1 score, while 0.5 / 500 was the second best. In task 2, 0.1 / 5000 had the highest score, followed by 0.1 / 2000 and 0.5 / 5000, and in task 3, 0.1 / 5000 obtained the highest score, while 0.001 / 5000 had the second best. In task 4, 0.1 / 2000 achieved the highest F1 score, followed by 0.1 / 5000 and 0.5 / 5000. In addition, given the same replacement rate, the maturity threshold value 500 had the lowest F1 scores in tasks 3 and 4. The lowest replacement rate of 0.001 also obtained the lowest F1 scores on task 4. 
	
	Table \ref{hplwf} displays the sensitivity analysis results of LwF. In task 1, 0.1 / 500 had the highest F1 score, followed by 0.001 / 2000 and 0.1 / 5000. In task 2, 0.5 / 2000 obtained the highest classification score, slightly higher than 0.001 / 500, 0.1 / 5000, and 0.001 / 5000. Moreover, in task 3, 0.1 / 2000 and 0.5 / 2000 acquired similar F1 scores, followed by 0.001 / 2000, and in task 4, 0.001 / 5000 had the best F1 score, slightly higher than 0.1 / 2000. Furthermore, for each parameter combination, the classification scores of LwF + CB were higher on some tasks. For instance, the LwF + CB with parameter combinations 0.001 / 2000, 0.001 / 5000, 0.1 / 2000, and 0.5 / 2000 obtained better F1 scores on task 4. While analyzing the effects of replacement rate and maturity threshold, given the same replacement rate, the maturity threshold value of 500 also had the lowest F1 scores in tasks 3 and 4. However, the lowest replacement rate 0.001 obtained relatively higher F1 scores on tasks 2 and 4.  
	
	Table \ref{hpewc} presents the hyperparameter tunning results of EWC with CB. In task 1, 0.5 / 500 acquired the best F1 score, which was nearly 3\% higher than that of the second best combination, i.e. 0.001 / 5000. In task 2, 0.5 / 2000 and 0.1 / 5000 obtained better scores than other combinations. In tasks 3 and 4, 0.001 / 5000 achieved the best weighted average F1 scores. However, EWC + CB failed to outperform Finetune + CB with most parameter combinations in most CIL tasks. Moreover, the best-performing combinations of EWC + CB in each task had lower classification scores than Finetune + CB. For instance, the highest F1 score that Finetune + CB can achieve is 11.72\% mean F1 score, while EWC + CB only gets 10.90\% average F1 score. Like Finetune + CB and LwF + CB, the maturity threshold value of 500 for EWC + CB also had the lowest classification results in tasks 3 and 4. The effects of replacement rate varied in different CIL tasks given the same maturity threshold values.
	
	Table \ref{hpreplay} shows the weighted average F1 scores of different parameter combinations of Replay with CB. In CIL tasks 1 to 4, 0.5 / 2000 had the highest scores, 1-5\% higher than the second-best combination. The standard deviation of 0.5 / 2000 was also lower than most other combinations on tasks 3 and 4. In addition, the classification scores were higher on each CIL task than Finetune + CB, LwF + CB, and EWC + CB, regardless of parameter combinations. The maturity threshold 2000 acquired better F1 scores in tasks 3 and 4 than most other parameter combinations, except for the replacement rate of 0.001 in task 4.
	
	Table \ref{hpicarl} presents the F1 scores for sensitivity analysis of iCarl with CB. In task 1, 0.5 / 500 and 0.001 / 5000 achieved $\geq$ 83\% mean weighted F1 scores and 0.001 / 5000 had the lowest standard deviation. In task 2, 0.5 / 2000 had the highest score, followed by 0.001 / 2000 and 0.1 / 5000. In task 3, 0.001 / 5000 and 0.1 / 500 obtained similar weighted F1 scores, while the standard deviation of combination 0.001 / 5000 was $\leq$2.00\%. In task 4, 0.001 / 5000 also acquired the best score with the lowest standard deviation. However, the parameter combinations of iCarl + CB with the highest F1 scores did not obtain better F1 scores on most tasks, except for task 4, compared to those of the Replay  + CB method. The effects of the replacement rate and maturity threshold for iCarl + CB differed in each CIL task. The parameter combinations with a maturity threshold value of 500 had lower classification results in tasks 3 and 4 than most other combinations, except those with a replacement rate of 0.001 in task 4 and 0.1 in task 3. 
	
	Table \ref{hpwa} displays the sensitivity analysis results of WA with CB. In task 1, 0.5 / 5000 had the highest F1 scores with the lowest standard deviation. In task 2, 0.1 / 5000 achieved the best classification result, while 0.1 / 2000 and 0.5 / 5000 obtained similar F1 scores. In tasks 3 and 4, 0.001 / 2000 obtained the best weighted F1 scores with lower standard deviation than most other parameter combinations. Like iCarl  + CB, WA + CB failed to outperform the Replay + CB method on tasks 2, 3, and 4 for most parameter combinations. The combinations with maturity threshold 500 also had lower F1 scores for tasks 3 and 4 in most experiments.   
	
	Table \ref{hpder} demonstrates the performance of different parameter combinations of DER + CB. In task 1, 0.001 / 500 and 0.5 / 5000 had the higher F1 score of $\geq$88\%, while the standard deviation of 0.5 / 5000 was about 2.45\%. In task 2, 0.1 / 2000 and 0.5 / 500 achieved $\geq$72\% classification scores, while the standard deviation of 0.1 / 2000 was much lower. In tasks 3 and 4, 0.1 / 2000 also had the highest weighted average F1 score, with a relatively low standard deviation on task 3 and the highest standard deviation on task 4. Furthermore, DER + CB achieved better classification scores than Replay + CB on tasks 1, 2, and 4 with most parameter combinations. The effects of the replacement rate and maturity threshold diversified concerning different parameter combinations in each CIL task; thus, the conclusion is that their effects on DER + CB are generally complex. The maturity threshold 500 had lower F1 scores in task 4 with replacement rates of 0.001 and 0.1, while the maturity threshold 5000 had lower F1 scores in task 3 with replacement rates of 0.001 and 0.1. 
	
	Table \ref{hpfoster} presents the sensitivity analysis results of FOSTER + CB. In task 1, 0.1 / 500 and 0.5 / 2000 achieved $\geq$97\% F1 scores with only around 1\% standard deviation. In task 2, 0.5 / 2000 also had the highest F1 score, while 0.5 / 2000 and 0.1 / 500 obtained $\geq$79\% scores on task 3. In task 4, 0.5 / 2000 acquired the highest weighted average F1 score, and 0.1 / 500 had the second-best score with a smaller standard deviation. Moreover, FOSTER + CB outperforms Replay + CB for most parameter combinations on all tasks. Like DER + CB, the effects of parameter combinations of FOSTER + CB also differed in each CIL task. For instance, given replacement rates 0.001 and 0.5, maturity threshold 500 obtained the lowest classification results in tasks 3 and 4, while it also achieved the best classification performance with replacement 0.1 in tasks 3 and 4. 
	
	Table \ref{hpmemo} demonstrates the classification results of different parameter combinations of MEMO + CB. In tasks 1, 0.001 / 5000 had the best F1 scores with relatively low standard deviations, followed by 0.001 / 2000. In task 2, 0.1 / 5000 obtained the best classification performance and 0.001 / 2000 also had the second-best result. In tasks 3 and 4, 0.001 / 2000 obtained the best classification scores. However, MEMO + CB underperforms Replay + CB on tasks 2 and 3, while it acquired a F1 score that was only slightly higher than that of Replay + CB on task 4 with hgiher standard deviation. In addition, a higher replacement rate can cause the performance degradation on tasks 3 and 4 for most parameter combinations.    
	
	In general, the parameter combinations of replacement rate and maturity threshold affected various CIL techniques differently in each CIL task. For some CIL methods, a low maturity threshold of 500 caused the degradation of classification performance in tasks 3 and 4. However, this phenomenon can not be observed for other CIL techniques. It is challenging to generate a conclusion about the effects of hyperparameter combinations on the performance of CIL + CB methods in different tasks. Due to the variations of different parameter combinations on different CIL tasks, the best replacement rate and maturity threshold combinations were decided based on the F1 scores of task 4, where the neural networks can access all the botanical origin classes at this stage. The selected parameter combinations for different CIL techniques with CB are shown in Table \ref{cbbest}.

	\begin{table}[ht]
		\centering
		\begin{tabular}{|c|c|c|}
			\hline
			\multicolumn{3}{|c|}{\textbf{Best hyperparameter of CB}} \\
			\hline
			\textbf{Method} & Replacement Rate & Maturity Threshold \\
			\hline
			Finetune + CB & 0.1 & 2000 \\
			\hline
			LwF + CB & 0.001 & 5000 \\
			\hline
			EWC + CB & 0.001 & 2000 \\
			\hline
			Replay + CB & 0.5 & 2000 \\
			\hline
			iCarl + CB & 0.001 & 5000 \\
			\hline
			WA + CB & 0.001 & 2000 \\
			\hline
			DER + CB & 0.1 & 2000 \\
			\hline
			FOSTER + CB & 0.5 & 2000 \\
			\hline
			MEMO + CB & 0.001 & 2000 \\
			\hline
		\end{tabular}
		\caption{Best combinations of CB's Replacement Rate and Maturity Threshold according to sensitivity analysis. }
		\label{cbbest}
	\end{table}
	
	\subsection{Performance Comparisons between CIL and CIL + CB}\label{cilcbcomp}
	According to the parameter combinations in Table \ref{cbbest}, the classification results of different CIL methods with CB are shown in Table \ref{tab2}. After comparing the classification results in Tables \ref{tab1} and \ref{tab2}, it seems that continuous backpropagation (CB) can be used to improve the class-incremental learning performance for most algorithms. The F1 scores of the three methods without exemplars increased after applying CB on task 4, while their performance fluctuated on tasks 1, 2, and 3: the score of Fintune + CB slightly inclined, while those of LwF + CB and EWC + CB declined on task 1; Fintune + CB and LwF + CB obtained higher F1 scores on task 2 than vanilla Fintune and LwF, while EWC + CB had a lower score with higher standard deviation; all the three algorithms with CB acquired lower F1 scores on task 3 compared to the corresponding vanilla CIL methods, while they all achieved higher F1 scores on task 4 with higher standard deviations. 
	
	For the algorithms using exemplars, the performance varied concerning both task and methods: in task 1, higher F1 scores were obtained by Replay(82.22\% to 83.86\%), iCarl (81.78\% to 83.24\%), WA (80.52\% to 81.58\%,), FOSTER (94.01\% to 97.20\%), and MEMO (76.06\% to 80.69\%), while lower scores were acquired by DER (89.04\% to 86.64\%). In task 2, classification performance can be improved after using CB for Replay (60.87\% to 66.18\%), DER (68.71\% to 72.34\%), FOSTER (57.09\% to 68.87\%), and MEMO (54.81\% to 60.31\%), while iCarl (58.91\% to 55.35\%) and WA (57.38\% to 55.53\%) experienced deterioration. On task 3, all CIL methods with CB had higher scores, including Replay (69.29\% to 73.20\%), iCarl (65.73\% to 68.13\%), WA (66.21\% to 69.00\%), DER (71.75\% to 73.13\%), FOSTER (75.30\% to 79.20\%), and MEMO (65.05\% to 68.31\%). In task 4, WA (68.91\% to 68.84\%) experienced performance decline after using Continual Backpropagation, while Replay (67.94\% to 70.11\%), iCarl (69.74\% to 71.89\%), DER (75.05\% to 75.73\%), FOSTER (71.24\% to 74.34\%), and and MEMO (63.51\% to 67.39\%)  achieved higher F1 scores by applying CB.  
	
	After combining Continual Backpropagation with other CIL algorithms, the performance deterioration was mainly observed by WA (on tasks 2 and 4). However, WA + CB had lower standard deviations on all the CIL tasks. The performance differences between vanilla WA and WA + CB were $\leq$2\% in task 2 with $\geq$7\% standard deviation and $\leq$0.1\% in task 4 with $\geq$3\% standard deviation. These deviations might not be significant, given the large standard deviation of their F1 scores. 
	
	Although DER, FOSTER, and MEMO are dynamic networks, FOSTER uses model compression to restrain the growth of model sizes, DER expands a new backbone for each new task, and MEMO uses the same generalized block for all tasks. Thus, most of the information in the newly expanded components for both algorithms could be more specific and useful for the corresponding tasks. Continuous backpropagation improves the network plasticity by reinitializing some of the neurons, which improves FOSTER performance more significantly than DER's. The reinitialization of neurons was more effective for FOSTER, which contains only two sub-networks, than DER, which includes one network backbone for each CIL task. For other algorithms, the neuronal reinitialization by CB can release neurons for newly encountered tasks. More experiments should be conducted to explore the reasons for this variation further. 
	
\begin{figure}[ht]
		\centering
		\includegraphics[scale=0.6]{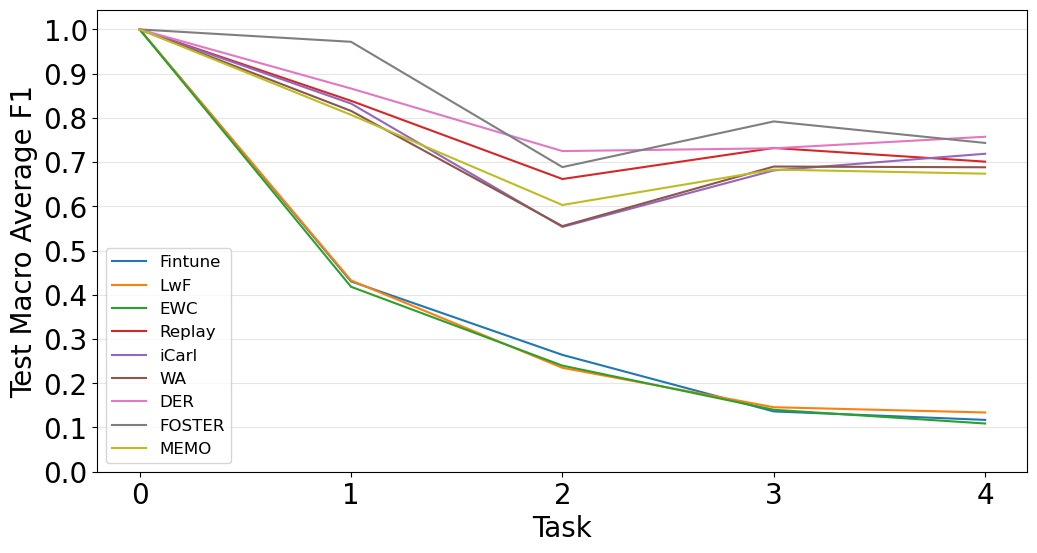}
		\caption{Comparison of different class-incremental learning algorithms with continual backpropagation. The mean values for the test weighted average F1 scores of the five rounds of experiments were shown for each class-incremental learning task. }
		\label{wcbf1}
	\end{figure}

	A similar pattern can also be observed from both Fig. \ref{f1} and Fig. \ref{wcbf1}: the classification accuracies' gaps between the algorithms using exemplars and the ones without exemplars were significant. Moreover, a performance increase can be achieved for most algorithms using exemplars from task 2 to task 3. In contrast, the methods without exemplars demonstrated declining classification scores as more botanical origin classes were added to the CIL task.
	
	\begin{table}[ht]
		
		\begin{center}
			\begin{tabular}{|c|c|c|c|c|c|}
				\hline
				\textbf{CIL}&\multicolumn{5}{c|}{\textbf{Weighted Average F1 Scores of Incremental Task No.}} \\
				\cline{2-6} 
				\textbf{Method} & \textbf{\textit{0}} & \textbf{\textit{1}} & \textbf{\textit{2}} & \textbf{\textit{3}} & \textbf{\textit{4}} \\
				\hline
				Finetune+CB & 100.00$\pm$0.00\% & 42.06$\pm$4.33\% & 26.70$\pm$2.10\% & 14.41$\pm$1.78\% & 11.37$\pm$1.82\% \\
				\cline{2-6}
				LwF+CB & 100.00$\pm$0.00\% & 43.25$\pm$6.92\% & 23.49$\pm$4.33\% & 14.58$\pm$2.30\% & 13.40$\pm$1.53\% \\
				\cline{2-6}
				EWC+CB & 100.00$\pm$0.00\% & 41.81$\pm$6.36\% & 23.99$\pm$3.97\% & 14.01$\pm$1.03\% & 10.90$\pm$2.67\% \\
				\cline{2-6}
				Replay+CB & 100.00$\pm$0.00\% & 83.86$\pm$4.81\% & 66.18$\pm$6.03\% & 73.20$\pm$2.04\% & 70.11$\pm$2.13\% \\
				\cline{2-6}
				iCarl+CB & 100.00$\pm$0.00\% & 83.24$\pm$1.67\% & 55.35$\pm$3.86\% & 68.13$\pm$1.98\% & 71.89$\pm$1.68\% \\
				\cline{2-6}
				WA+CB & 100.00$\pm$0.00\% & 81.58$\pm$3.19\% & 55.53$\pm$7.75\% & 69.00$\pm$2.83\% & 68.84$\pm$3.42\% \\
				\cline{2-6}
				DER+CB & 100.00$\pm$0.00\% & 86.64$\pm$2.90\% & 72.50$\pm$1.85\% & 73.13$\pm$2.08\% & 75.73$\pm$5.48\% \\
				\cline{2-6}
				FOSTER+CB & 100.00$\pm$0.00\% & 97.20$\pm$0.99\% & 68.87$\pm$6.33\% & 79.20$\pm$3.96\% & 74.34$\pm$3.51\% \\
				\cline{2-6}
				MEMO+CB & 100.00$\pm$0.00\% & 82.31$\pm$2.94\% & 60.42$\pm$5.41\% & 71.54$\pm$1.51\% & 70.94$\pm$3.13\% \\
				\hline
			\end{tabular}
			\caption{Incremental Learning with Continual Backpropagation Classification Results}
			\label{tab2}
		\end{center}
	\end{table}

	\subsection{Kernel Weight Distribution Analysis}\label{cilkwdanalysis}
	
	The statistical descriptions and kernel weight distributions of the last convolutional layer are also analyzed to evaluate the effects of CB. The mean and standard deviation of the last convolutional layer from different CIL methods and CIL + CB methods are shown in Tables \ref{table:meanstdcil} and \ref{table:meanstdcilcb}, while the skewness and kurtosis of the kernel weight distributions for the last convolutional layers of different vanilla CIL methods and CIL + CB are displayed in Tables \ref{table:skkcil} and \ref{table:skkcilcb}. Moreover, differet CIL and CIL + CB techniques' kernel density estimations of the last layers on each CIL task are presented in Appendix Chapter \ref{cilkde}.
	
	\subsubsection{Mean and Standard Deviation}
	
\begin{figure}[ht]
		\centering
		\includegraphics[scale=0.6]{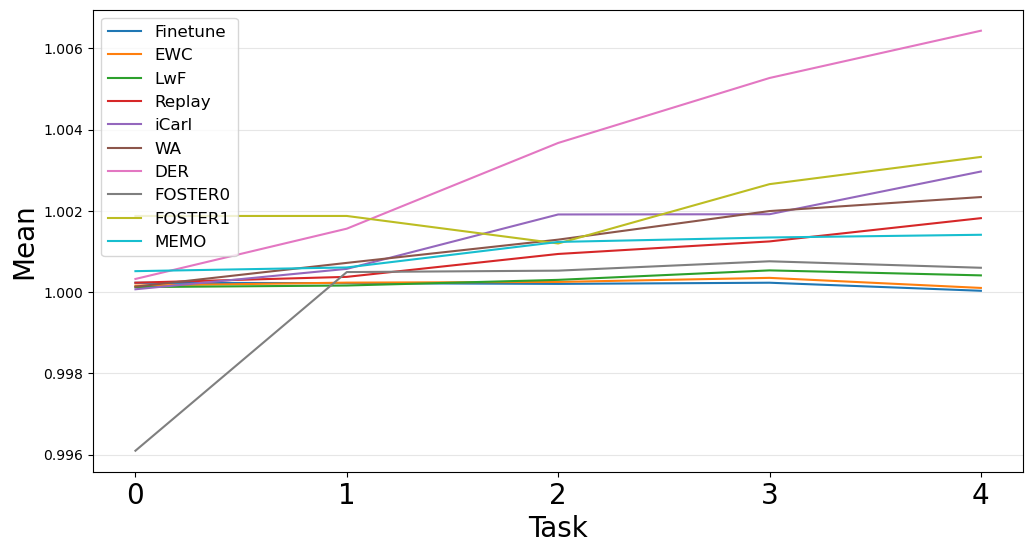}
		\caption{Mean value of Last Layer's Weights for CIL Methods on Each Task.}
		\label{cilmean}
	\end{figure}

\begin{figure}[ht]
		\centering
		\includegraphics[scale=0.6]{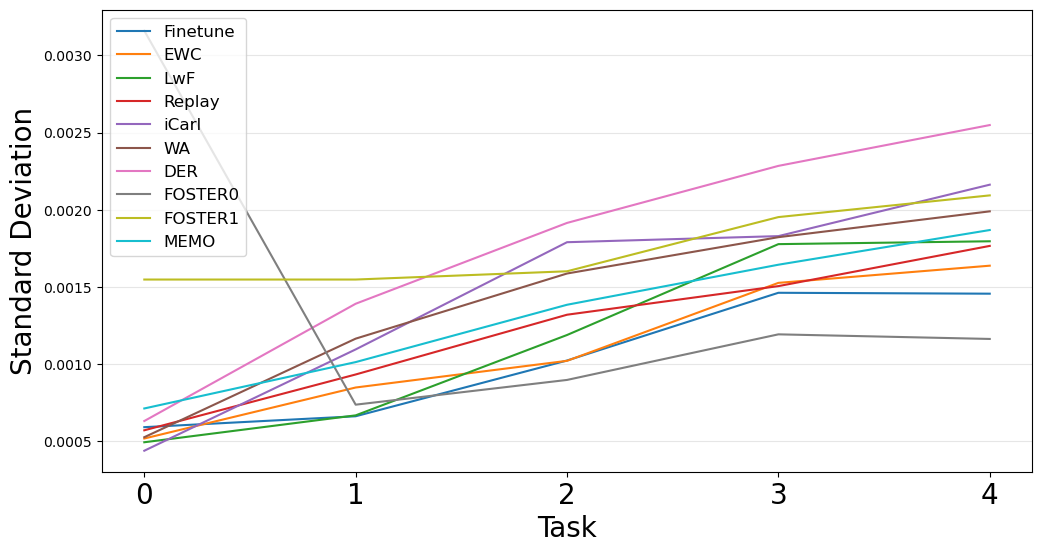}
		\caption{Standard Deviation value of Last Layer's Weights for CIL Methods on Each Task.}
		\label{cilstd}
	\end{figure}

	Table \ref{table:meanstdcil} presents the mean and standard deviations of the last convolutional layers' weights for different CIL techniques. The mean values and standard deviations are displayed in Figs \ref{cilmean} and \ref{cilstd}. For most CIL methods, the mean values of the last layer weights incline when the networks are exposed to more honey botanical origin classes. For the three CIL algorithms without exemplars, the mean values increase from task 0 to 3 while they decline on task 4. In contrast, the mean values of the layer weights for the CIL methods using the exemplar set continue to grow as more CIL tasks are conducted. In addition, the layer weights' mean values of CIL techniques with exemplar sets are generally higher in most tasks. For example, the lowest layer weight's mean value of CIL methods with exemplars in task 1 is 1.000445 from FOSTER's teacher network, while the highest layer weight's mean value of CIL without exemplars is 1.00327 from EWC. Fig \ref{cilmean} presents this apparent trend. Moreover, Table \ref{tab1} shows that DER and FOSTER achieved higher weighted average F1 scores in the experiments, and Table \ref{table:meanstdcil} demonstrates that DER and FOSTER's student network also had the highest mean values in almost all tasks.
	
	The layer weights' standard deviations of all CIL methods increase as more CIL tasks are trained. For instance, the layer weights' standard deviation of LwF for each CIL task is 0.000534, 0.000708, 0.001154, 0.001570, and 0.001831, and that of iCarl for each CIL task is 0.000646, 0.001233, 0.001690, 0.001836, and 0.002064. This trend is evident in Fig \ref{cilstd}. Like the layer weights' mean values, the layer weights' standard deviation of CIL methods with exemplars are higher than those without exemplars in most CIL tasks. For instance, except for FOSTER's teacher network, MEMO had the lowest layer weights' standard deviation among the CIL methods with exemplars in task 1. However, this value is still higher than the layer weights' standard deviation of EWC, which had the highest values among CIL methods without exemplars. Furthermore, DER and FOSTER's student network also attained higher layer weights' standard deviations in tasks 3 and 4. Table \ref{tab1} shows that DER and FOSTER also achieved higher weighted average F1 scores in these two tasks. This observation is consistent with the definition of incremental learning: the values of the neural networks' layer weights in CIL tasks become more diversified to improve the representativeness as more classes are known. 
	
\begin{figure}[ht]
		\centering
		\includegraphics[scale=0.6]{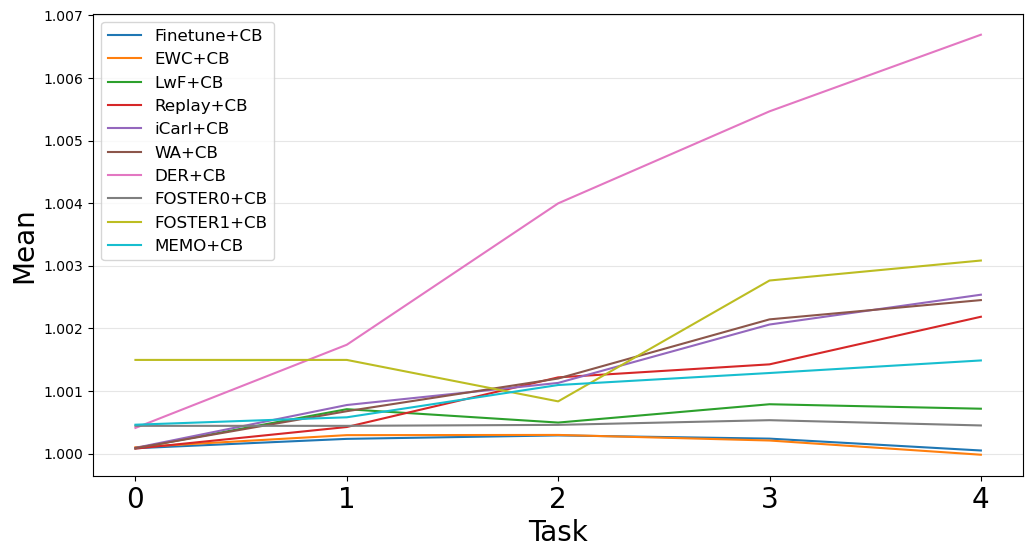}
		\caption{Mean value of Last Layer's Weights for CIL + CB Methods on Each Task.}
		\label{cilmeancb}
	\end{figure}

\begin{figure}[ht]
		\centering
		\includegraphics[scale=0.6]{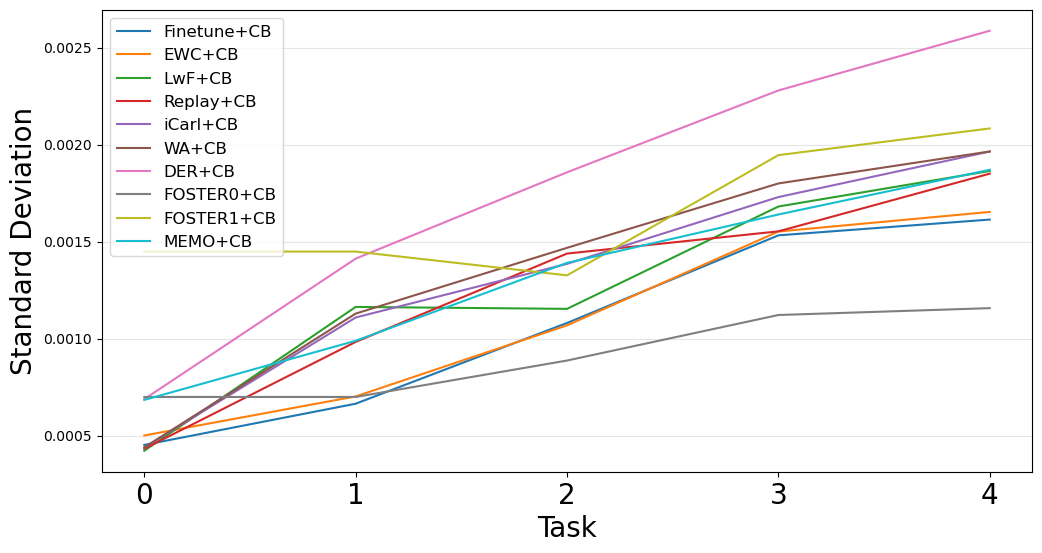}
		\caption{Standard Deviation value of Last Layer's Weights for CIL + CB Methods on Each Task.}
		\label{cilstdcb}
	\end{figure}

	Table \ref{table:meanstdcilcb} exhibits the last layer weights' mean and standard deviation of CIL methods with CB in all tasks. Figs \ref{cilmeancb} and \ref{cilstdcb} demonstrate that the mean and standard deviation values of layer weights for CIL + CB methods also generally increase as the networks are trained for more tasks. Moreover, Table \ref{table:meanstdcilcb} presents that the CIL + CB techniques with exemplars have higher layer weights' mean and standard deviation values than those without exemplars. However, while comparing the values in Tables \ref{table:meanstdcil} and \ref{table:meanstdcilcb}, the classification scores' improvements in Tables \ref{tab1} and \ref{tab2} are not always related to higher mean and standard deviation values of layer weights. For example, DER and DER + CB acquired mean values of 1.005731 and 1.005467 in task 3 and 1.006603 and 1.006691 in task 4, respectively. The standard deviation values of DER and DER + CB are 0.002190 and 0.002279 in Task 3 and 0.002530 and 0.002587 in Task 4. In addition, the standard deviations of the student networks' layer weights from FOSTER and FOSTER + CB are 0.001969 and 0.001946 in task 3 and 0.002100 and 0.002083 in task 4, respectively. However, DER + CB achieved higher classification results than DER, and FOSTER + CB had better weighted F1 scores than FOSTER in both tasks.
	
	Generally speaking, layer weights' mean and standard deviation values increase as more CIl tasks are conducted. The CIL methods using exemplars have generally higher mean and standard deviation values of layer weights than CIL methods without exemplars. These phenomena can be observed in vanilla CIL techniques and CIL + CB methods. However, the classification performance improvements by applying CB are not positively related to layer weights' mean and standard deviation values. The reinitialization of less-used neurons using CB decreases the layer weight's standard deviation values while improving the classification performance for most CIL methods. 
	
	\begin{table}[ht]
		\centering
		\begin{tabular}{ |c|c|c|c|c|c|c| }
			\hline
			\multicolumn{7}{|c|}{Last Convolutional Layer Kernel Weights Description of Task No.} \\
			\hline
			Values & Method & 0 & 1 & 2 & 3 & 4  \\
			\hline
			\multirow{9}{*}{Mean} 
			& Finetune & 1.000076 & 1.000242 & 1.000264 & 1.000206 & 1.000043 \\
			\cline{3-7}
			& LwF & 1.000127 & 1.000205 & 1.000400 & 1.000718 & 1.000541 \\
			\cline{3-7}
			& EWC & 1.000148 & 1.000327 & 1.000226 & 1.000298 & 1.000101 \\
			\cline{3-7}
			& Replay & 1.000284 & 1.000520 & 1.000970 & 1.001626 & 1.002244 \\
			\cline{3-7}
			& iCarl & 1.000332 & 1.000810 & 1.001764 & 1.002155 & 1.002608 \\
			\cline{3-7}
			& WA & 1.000079 & 1.000823 & 1.001316 & 1.002393 & 1.002529 \\
			\cline{3-7}
			& DER & 1.000325 & 1.001563 & 1.003672 & 1.005270 & 1.006603 \\
			\cline{3-7}
			& FOSTER t & 0.996098 & 1.000445 & 1.000482 & 1.000694 & 1.000528 \\
			\cline{3-7}
			& FOSTER s & - & 1.001875 & 1.001198 & 1.002658 & 1.003328 \\
			\cline{3-7}
			& MEMO & 1.000517 & 1.000608 & 1.001236 & 1.001346 & 1.001412 \\
			\hline
			\hline
			\multirow{4}{*}{Standard} 
			& Finetune & 0.000437 & 0.000927 & 0.001044 & 0.001475 & 0.001627 \\
			\cline{3-7}
			& LwF & 0.000534 & 0.000708 & 0.001154 & 0.001570 & 0.001831 \\
			\cline{3-7}
			& EWC & 0.000542 & 0.000969 & 0.001035 & 0.001491 & 0.001655 \\
			\cline{3-7}
			& Replay & 0.000614 & 0.001018 & 0.001358 & 0.001611 & 0.001920 \\
			\cline{3-7}
			& iCarl & 0.000646 & 0.001233 & 0.001690 & 0.001836 & 0.002064 \\
			\cline{3-7}
			\multirow{4}{*}{Deviation}
			& WA & 0.000439 & 0.001175 & 0.001501 & 0.001830 & 0.002033 \\
			\cline{3-7}
			& DER & 0.000630 & 0.001391 & 0.001915 & 0.002284 & 0.002530 \\
			\cline{3-7}
			& FOSTER t & 0.003160 & 0.000707 & 0.000904 & 0.001145 & 0.001165 \\
			\cline{3-7}
			& FOSTER s & - & 0.001548 & 0.001601 & 0.001952 & 0.002093 \\
			\cline{3-7}
			& MEMO & 0.000712 & 0.001012 & 0.001384 & 0.001644 & 0.001868 \\
			\hline
		\end{tabular}
		\caption{Mean and Standard Deviation of Last Convolutional Layer's Kernel Weights for different CIL techniques. }
		\label{table:meanstdcil}
	\end{table}
	
	\begin{table}[ht]
		\centering
		\begin{tabular}{ |c|c|c|c|c|c|c| }
			\hline
			\multicolumn{7}{|c|}{Last Convolutional Layer Kernel Weights Description of Task No.} \\
			\hline
			Values & Method & 0 & 1 & 2 & 3 & 4  \\
			\hline
			\multirow{9}{*}{Mean} 
			& Finetune + CB & 1.000087 & 1.000237 & 1.000292 & 1.000240 & 1.000051 \\
			\cline{3-7}
			& LwF + CB & 1.000086 & 1.000710 & 1.000497 & 1.000707 & 1.000718 \\
			\cline{3-7}
			& EWC + CB & 1.000103 & 1.000296 & 1.000299 & 1.000210 & 0.999983 \\
			\cline{3-7}
			& Replay + CB & 1.000081 & 1.000426 & 1.001219 & 1.001427 & 1.002187 \\
			\cline{3-7}
			& iCarl + CB & 1.000093 & 1.000777 & 1.001130 & 1.002063 & 1.002539 \\
			\cline{3-7}
			& WA + CB & 1.000089 & 1.000678 & 1.001200 & 1.002145 & 1.002453 \\
			\cline{3-7}
			& DER + CB & 1.000411 & 1.001740 & 1.003997 & 1.005467 & 1.006691 \\
			\cline{3-7}
			& FOSTER t + CB & 1.000445 & 1.000445 & 1.000460 & 1.000535 & 1.000451 \\
			\cline{3-7}
			& FOSTER s + CB & - & 1.001497 & 1.000835 & 1.002765 & 1.003085 \\
			\cline{3-7}
			& MEMO + CB & 1.000463 & 1.000581 & 1.001097 & 1.001288 & 1.001489 \\
			\hline
			\hline
			\multirow{5}{*}{Standard} 
			& Finetune + CB & 0.000453 & 0.000665 & 0.001081 & 0.001532 & 0.001614 \\
			\cline{3-7}
			& LwF + CB & 0.000423 & 0.001164 & 0.001154 & 0.001742 & 0.001864 \\
			\cline{3-7}
			& EWC + CB & 0.000501 & 0.000702 & 0.001069 & 0.001552 & 0.001653 \\
			\cline{3-7}
			& Replay + CB & 0.000433 & 0.000983 & 0.001439 & 0.001553 & 0.001850 \\
			\cline{3-7}
			& iCarl + CB & 0.000439 & 0.001109 & 0.001386 & 0.001730 & 0.001964 \\
			\cline{3-7}
			\multirow{4}{*}{Deviation}
			& WA + CB & 0.000440 & 0.001129 & 0.001468 & 0.001800 & 0.001965 \\
			\cline{3-7}
			& DER + CB & 0.000688 & 0.001412 & 0.001857 & 0.002279 & 0.002587 \\
			\cline{3-7}
			& FOSTER t + CB & 0.000700 & 0.000700 & 0.000887 & 0.001122 & 0.001157 \\
			\cline{3-7}
			& FOSTER s + CB & - & 0.001449 & 0.001326 & 0.001946 & 0.002083 \\
			\cline{3-7}
			& MEMO + CB & 0.000684 & 0.000989 & 0.001391 & 0.001640 & 0.001870 \\
			\hline
		\end{tabular}
		\caption{Mean and Standard Deviation of Last Convolutional Layer's Kernel Weights for different CIL techniques with CB. }
		\label{table:meanstdcilcb}
	\end{table}
	
	\subsubsection{Skewness and Kurtosis}
	
	The \textbf{skewness} is designed to measure the level of asymmetry of a distribution: the positive skewness means that the values in the distribution are more concentrated on the right side; negative skewness indicates longer or fatter left tail of the distribution than the right tail; the skewness with a value of zero indicates that the distribution is symmetrical \citep{skewness1998}. For a positively skewed distribution, the mean is greater than the median and the mode (Mean > Median > Mode). For a negatively skewed distribution, the mode is greater than both the mean and the median (Mean < Median < Mode). 
	
	The \textbf{kurtosis} is developed to assess a distribution's peakedness, tailedness, or flatness, where the tailedness indicates the frequency of outliers in a distribution \citep{kurtosis1988}. A distribution with kurtosis of 3 is a normal distribution, i.e., Mesokurtic. A kurtosis greater than 3 means that the distribution has sharp peaks and thick tails (more frequent outliers), i.e., Leptokurtic. In contrast, a kurtosis less than 3 means that the distribution has a flat peak and thin tails (less frequent outliers), i.e., Platykurtic. In this work, the Kurtosis values are calculated using Fisher's definition of kurtosis, where the kurtosis of normal distribution is 0.
	
	The layer weights' distributions can be visualized with kernel density estimation graphs. The kernel density estimations of the last layers on each CIL task are presented in Fig. \ref{kft} for Finetune, Fig. \ref{kftcb} for Finetune + CB, Fig. \ref{klwf} for LwF, Fig. \ref{klwfcb} for LwF + CB, Fig. \ref{kewc} for EWC, Fig. \ref{kewccb} for EWC + CB, Fig. \ref{kreplay} for Replay, Fig. \ref{kreplaycb} for Replay + CB, Fig. \ref{kicarl} for iCarl, Fig. \ref{kicarlcb} for iCarl + CB, Fig. \ref{kwa} for WA, Fig. \ref{kwacb} for WA + CB, Fig. \ref{kder} for DER, Fig. \ref{kdercb} for DER + CB, Fig. \ref{kfoster0} for the teacher network of FOSTER, Fig. \ref{kfoster0cb} for teacher network of FOSTER + CB, Fig. \ref{kfoster1} for student network of FOSTER, Fig. \ref{kfoster1cb} for student network of FOSTER + CB, Fig. \ref{kmemo0} for the first task-specific adaptive extractor of MEMO, and Fig. \ref{kmemo0cb} for the first task-specific adaptive extractor of MEMO + CB. The kernel density estimation of the first task-specific adaptive extractor for MEMO and MEMO + CB were shown because this one existed in all CIL tasks, while others were only added to the network in the corresponding incremental tasks. 
	
\begin{figure}[ht]
		\centering
		\includegraphics[scale=0.6]{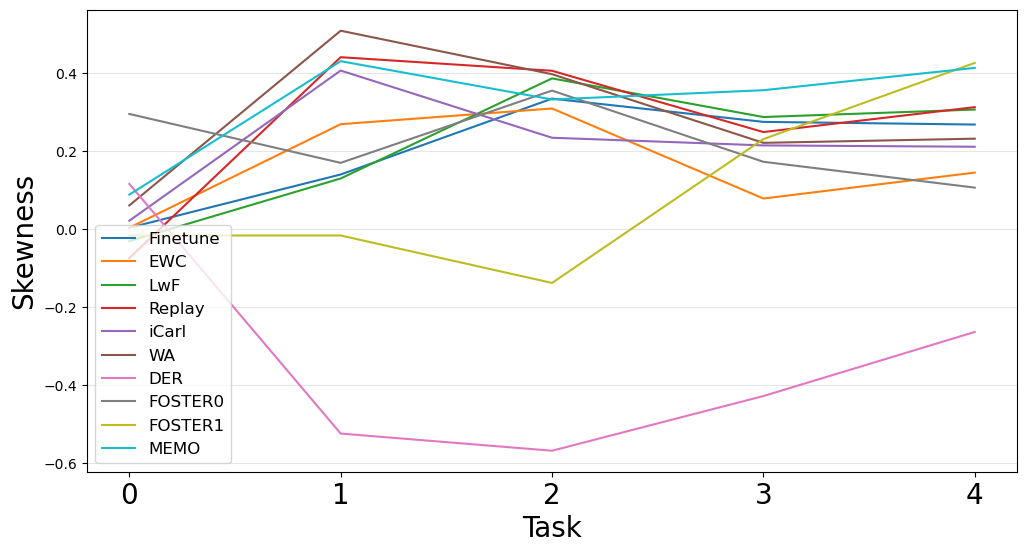}
		\caption{Skewness of Last Layer's Weights for CIL Methods on Each Task.}
		\label{cilskew}
	\end{figure}

\begin{figure}[ht]
		\centering
		\includegraphics[scale=0.6]{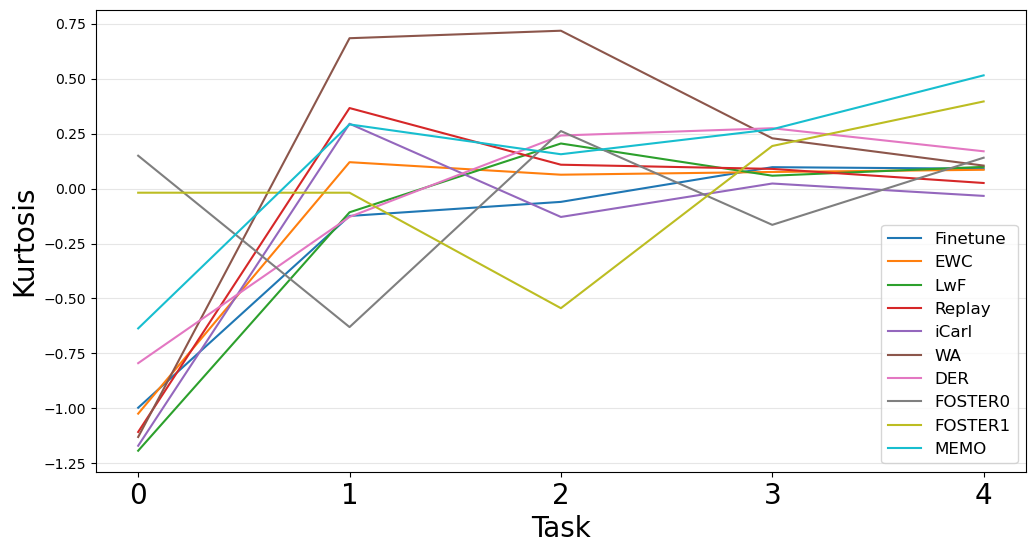}
		\caption{Kurtosis of Last Layer's Weights for CIL Methods on Each Task.}
		\label{cilkurt}
	\end{figure}

	Table \ref{table:skkcil} displays each task's last layer weights' skewness and kurtosis values of different CIL methods. Figs \ref{cilskew} and \ref{cilkurt} are graphs of different CIL methods' layer weights in each task. The skewness values of some CIL methods, including Finetune, LwF, and EWC, increase from task 0 to task 2, then decrease from task 3 to task 4. The last layer weights' skewness values of other methods, such as iCarl and WA, increase from task 0 to task 1, then decline from task 2 to task 4. The three dynamic neural networks have distinct skewness values in the CIL tasks: the skewness of DER's last layer weight is significantly lower than all the other CIL methods and is negative in all tasks; the skewness values of MEMO and FOSTER student network's last layer weights decrease from task 1 to 2, then they significantly increase from task 2 to 4 and have the highest skewnesses in task 4 among all CIL methods. The increasing skewness indicates that the values of neurons in layer weights become more concentrated on the right side of the distributions as more CIL tasks.
	
	The layer weights' kurtosis varied significantly for different CIL techniques. For instance, the kurtosis values of layer weights for Finetune, LwF, EWC, and iCarl fluctuated within different CIL tasks. In contrast, the kurtosis of WA and FOSTER student network's layer weights dropped from task 1 to task 2, then increased from task 2 to 4. The positive kurtosis values represent sharp peaks and thicker tails (more outliers). The last layer weights of CIL methods using exemplars have generally larger kurtosis values than those of the CIL methods without exemplars. Moreover, the sharp drop in the kurtosis value of DER from task 3 to task 4 may indicate that the trained network considers the honey botanical origins in task 4 similar to previously seen classes. Thus, the values of neurons in the layer weights become less diversified and have thinner tails. However, most of the other CIL methods all experience inclines of kurtosis values from task 3 to task 4. Therefore, this observation cannot explain the confusion of Mānuka honey with Rewarewa and Tawari honey types in Appendix Chapter \ref{cilconfusion}. Table \ref{cildata} demonstrates that Rewarewa and Tawari honey types are accessible in task 4. 
	
\begin{figure}[ht]
		\centering
		\includegraphics[scale=0.6]{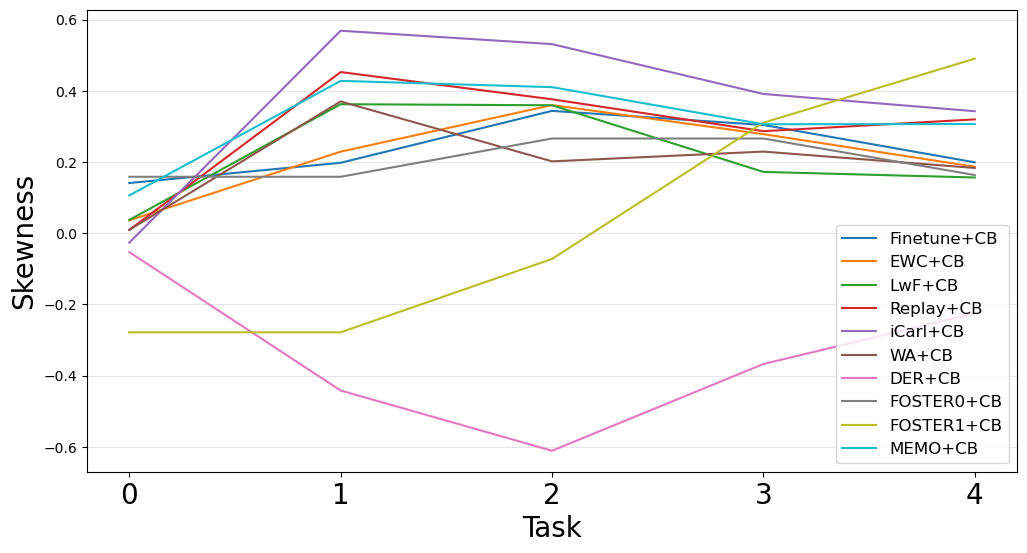}
		\caption{Skewness of Last Layer's Weights for CIL + CB Methods on Each Task.}
		\label{cilskewcb}
	\end{figure}

\begin{figure}[ht]
		\centering
		\includegraphics[scale=0.6]{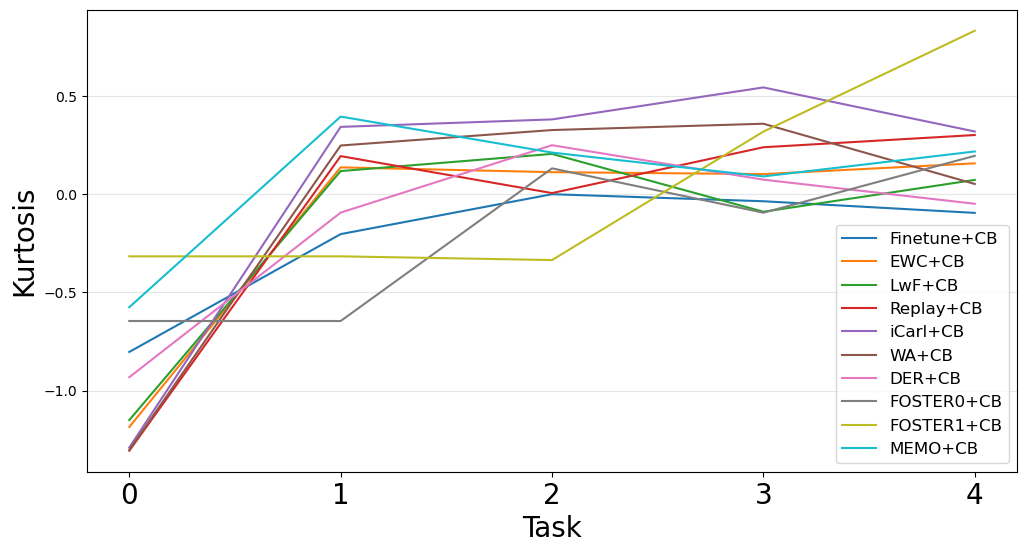}
		\caption{Kurtosis of Last Layer's Weights for CIL + CB Methods on Each Task.}
		\label{cilkurtcb}
	\end{figure}

	Table \ref{table:skkcilcb} presents the last layer weights' skewness and kurtosis of CIL + CB techniques. Figs \ref{cilskewcb} and \ref{cilkurtcb} plot line graphs to exhibit the changing of skewness and kurtosis values for CIL + CB methods in different CIL tasks. Fig \ref{cilskewcb} shows that the skewness values of most CIL + CB methods decrease after task 1, except for FOSTER + CB's student network and DER + CB. In contrast, Fig \ref{cilkurtcb} demonstrates that most CIL + CB methods' last layer weights' kurtosis values are positive after task 1. And the kurtosis of the FOSTER student's last layer weight continues to increase after task 2. This trend indicates that the last layer weight's distributions of different CIL + CB techniques have thick tails, i.e., frequent outliers, to keep the representativeness of the incrementally trained neural networks.

	\begin{table}[ht]
		\centering
		\begin{tabular}{ |c|c|c|c|c|c|c| }
			\hline
			\multicolumn{7}{|c|}{Last Convolutional Layer Kernel Weights Description of Task No.} \\
			\hline
			Values & Method & 0 & 1 & 2 & 3 & 4 \\
			\hline
			\multirow{9}{*}{Skewness} 
			& Finetune & -0.000717 & 0.265106 & 0.352097 & 0.213042 & 0.18111 \\
			\cline{3-7}
			& LwF & -0.064162 & 0.044047 & 0.361522 & 0.218138 & 0.15288 \\
			\cline{3-7}
			& EWC & 0.010717 & 0.43067 & 0.492257 & 0.1802 & 0.142189 \\
			\cline{3-7}
			& Replay & -0.076671 & 0.391697 & 0.338115 & 0.28272 & 0.328623 \\
			\cline{3-7}
			& iCarl & 0.12775 & 0.504157 & 0.379194 & 0.306965 & 0.285586 \\
			\cline{3-7}
			& WA & 0.023165 & 0.576929 & 0.399448 & 0.235613 & 0.255741 \\
			\cline{3-7}
			& DER & 0.11665 & -0.523995 & -0.568029 & -0.427707 & -0.26329 \\
			\cline{3-7}
			& FOSTER t & 0.296183 & 0.047392 & 0.257141 & 0.208529 & 0.089259 \\
			\cline{3-7}
			& FOSTER s & - & -0.015747 & -0.137493 & 0.231634 & 0.427099 \\
			\cline{3-7}
			& MEMO & 0.08901 & 0.43158 & 0.333275 & 0.357083 & 0.414311 \\
			\hline
			\hline
			\multirow{9}{*}{Kurtosis} 
			& Finetune & -1.236336 & -0.088612 & 0.035211 & -0.215863 & -0.140725 \\
			\cline{3-7}
			& LwF & -1.408444 & -0.399401 & 0.1227 & 0.047949 & 0.175021 \\
			\cline{3-7}
			& EWC & -1.300493 & 0.219906 & 0.458549 & -0.11564 & -0.052724 \\
			\cline{3-7}
			& Replay & -1.04532 & 0.18449 & 0.191213 & 0.284048 & 0.314298 \\
			\cline{3-7}
			& iCarl & -0.744155 & 0.254305 & 0.140389 & 0.269831 & 0.086415 \\
			\cline{3-7}
			& WA & -1.204841 & 0.446765 & 0.21086 & 0.103067 & 0.200762 \\
			\cline{3-7}
			& DER & -0.794467 & -0.128997 & 0.241315 & 0.274526 & 0.169552 \\
			\cline{3-7}
			& FOSTER t & 0.149461 & -0.894724 & -0.139975 & 0.015904 & 0.045179 \\
			\cline{3-7}
			& FOSTER s & - & -0.01887 & -0.544334 & 0.193651 & 0.396319 \\
			\cline{3-7}
			& MEMO & -0.636295 & 0.29189 & 0.156126 & 0.269694 & 0.515061 \\
			\hline
		\end{tabular}
		\caption{Skewness and Kurtosis of Last Convolutional Layer's Kernel Weights for different CIL techniques.}
		\label{table:skkcil}
	\end{table}
	
	\begin{table}[ht]
		\centering
		\begin{tabular}{ |c|c|c|c|c|c|c| }
			\hline
			\multicolumn{7}{|c|}{Last Convolutional Layer Kernel Weights Description of Task No.} \\
			\hline
			Values & Method & 0 & 1 & 2 & 3 & 4 \\
			\hline
			\multirow{9}{*}{Skewness} 
			& Finetune + CB & 0.140951 & 0.197724 & 0.343533 & 0.303885 & 0.199035 \\
			\cline{3-7}
			& LwF + CB & 0.037254 & 0.362197 & 0.359239 & 0.172105 & 0.156444 \\
			\cline{3-7}
			& EWC + CB & 0.036225 & 0.228959 & 0.359787 & 0.278157 & 0.18671 \\
			\cline{3-7}
			& Replay + CB & 0.009183 & 0.452464 & 0.376064 & 0.286482 & 0.319602 \\
			\cline{3-7}
			& iCarl + CB & -0.026435 & 0.568583 & 0.530934 & 0.390976 & 0.342551 \\
			\cline{3-7}
			& WA + CB & 0.00959 & 0.370035 & 0.201724 & 0.229192 & 0.183187 \\
			\cline{3-7}
			& DER + CB & -0.053126 & -0.441742 & -0.6108 & -0.367307 & -0.224847 \\
			\cline{3-7}
			& FOSTER t + CB & 0.158471 & 0.158471 & 0.26591 & 0.265903 & 0.163389 \\
			\cline{3-7}
			& FOSTER s + CB & - & -0.278318 & -0.072293 & 0.31034 & 0.49007 \\
			\cline{3-7}
			& MEMO + CB & 0.106058 & 0.427773 & 0.410126 & 0.30603 & 0.306594 \\
			\hline
			\hline
			\multirow{9}{*}{Kurtosis} 
			& Finetune + CB & -0.802685 & -0.202812 & 0.000349 & -0.03553 & -0.094773 \\
			\cline{3-7}
			& LwF + CB & -1.14957 & 0.118085 & 0.205561 & -0.088902 & 0.073125 \\
			\cline{3-7}
			& EWC + CB & -1.184759 & 0.13709 & 0.112582 & 0.102865 & 0.157359 \\
			\cline{3-7}
			& Replay + CB & -1.305648 & 0.194304 & 0.006052 & 0.239401 & 0.30169 \\
			\cline{3-7}
			& iCarl + CB & -1.289511 & 0.343121 & 0.381183 & 0.543909 & 0.319636 \\
			\cline{3-7}
			& WA + CB & -1.301758 & 0.248261 & 0.32698 & 0.359167 & 0.052479 \\
			\cline{3-7}
			& DER + CB & -0.931226 & -0.093256 & 0.249794 & 0.074408 & -0.048344 \\
			\cline{3-7}
			& FOSTER t + CB & -0.645224 & -0.645224 & 0.131961 & -0.093733 & 0.195678 \\
			\cline{3-7}
			& FOSTER s + CB & - & -0.315992 & -0.334915 & 0.319105 & 0.832708 \\
			\cline{3-7}
			& MEMO + CB & -0.57514 & 0.395608 & 0.212321 & 0.090933 & 0.21788 \\
			\hline
		\end{tabular}
		\caption{Skewness and Kurtosis of Last Convolutional Layer's Kernel Weights for different CIL techniques with CB.}
		\label{table:skkcilcb}
	\end{table}
	
	\section{Conclusion}
	
	In this research, the class-incremental learning techniques were applied to the honey hyperspectral imaging dataset. A variety of CIL algorithms, as described in Section \ref{cilmethod}, were examined using the honey HSI dataset. The experiments show that exemplars from previous classes can significantly improve classification results, as displayed in Section \ref{cilresult}. Moreover, the continuous backpropagation algorithm was proposed to combine with other CIL methods, as introduced in Section \ref{continualbackprop}. Most algorithms experienced performance improvement after applying CB, as shown in Tables \ref{tab1} and \ref{tab2}. The neural networks in CIL tasks can attain more plasticity to improve the classification performance by reinitializing a small fraction of less-used neurons. The kernel weight distributions of different CIL and CIL with CB methods' the last layer were analyzed with statistical descriptions, skewness, and kurtosis. The incrementally trained neural networks positively relate to the layer weights' mean and standard deviations. As more incremental training tasks are conducted, the mean and standard deviations of the last layer weights generally increase linearly. The peaks of the layer weights also shift with the increasing number of CIL tasks, and distributions of the neural network weights have thick tails to preserve the representativeness for newly encountered classes. 
	
	However, gaps exist between the fully retrained networks and the incrementally trained ones on the incrementally learning tasks. In addition, the performance variations between CIL methods with exemplar sets and the ones without exemplar sets are significant. A promising but challenging future research direction is improving CIL techniques' performance without using an exemplar set. The other direction for future research is to improve the performance of CIL methods using exemplar sets, whose main challenge is the imbalanced training dataset due to the limited number of exemplars. Moreover, different CIL techniques represent varying performance on different CIL tasks; most algorithms do not consistently outperform or underperform other methods. An assumption about the performance degradation on specific CIL models after combining CB is briefly discussed in Section \ref{resultcil}. Further investigation should be conducted to analyze this phenomenon to understand better the mechanisms of CIL techniques on the hyperspectral imaging dataset.
	\\
	\\
	\textit{CRediT authorship contribution statement}\\
	\textbf{Guyang Zhang}: Conceptualization, Methodology, Software, Investigation, Writing- Original draft, Visualization, Writing- Reviewing and Editing. \textbf{Waleed Abdulla}: Conceptualization, Project administration, Resources, Visualization, Supervision, Writing- Reviewing and Editing.


\newpage

	\appendix
	\counterwithin{figure}{section}
	\counterwithin{table}{section}
		
	\section{\textbf{\Large Confusion Matrix of CIL Methods}}\label{cilconfusion}
	
	\begin{figure}[H]
		\centering
		\includegraphics[scale=0.7]{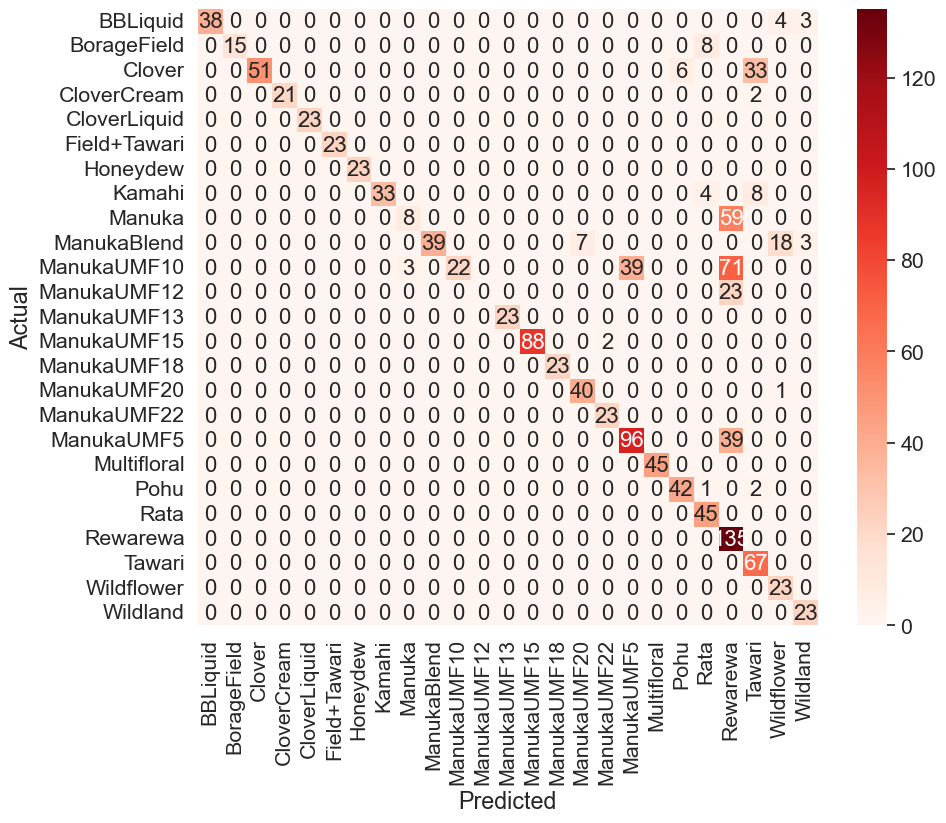}
		\caption{Confusion matrix of DER on Task 4.}
		\label{der44}
	\end{figure}
	
	\begin{figure}[H]
		\centering
		\includegraphics[scale=0.7]{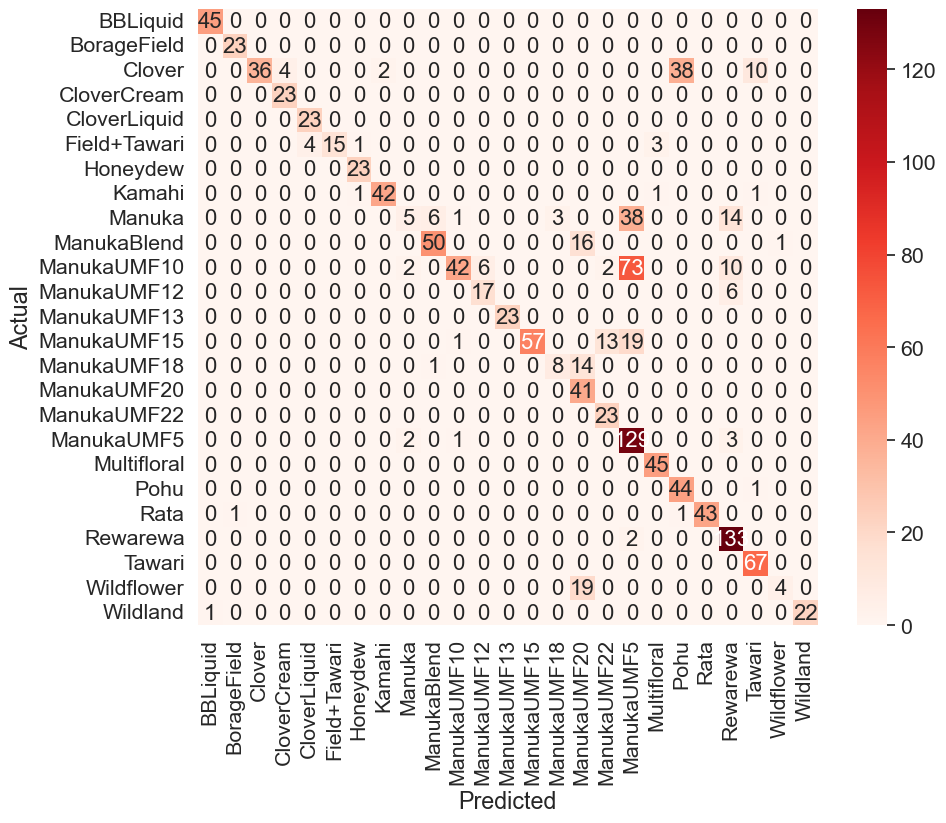}
		\caption{Confusion matrix of FOSTER on Task 4.}
		\label{foster44}
	\end{figure}
	
	\begin{figure}[H]
		\centering
		\includegraphics[scale=0.7]{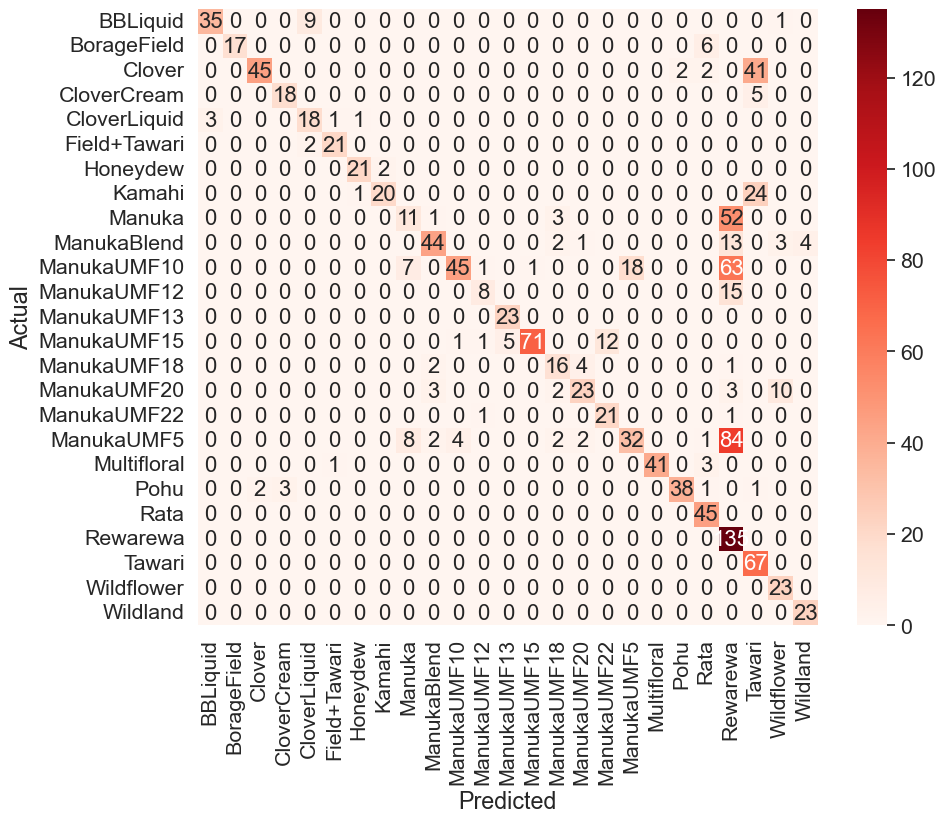}
		\caption{Confusion matrix of MEMO on Task 4.}
		\label{memo44}
	\end{figure}
	
	\begin{figure}[H]
		\centering
		\includegraphics[scale=0.7]{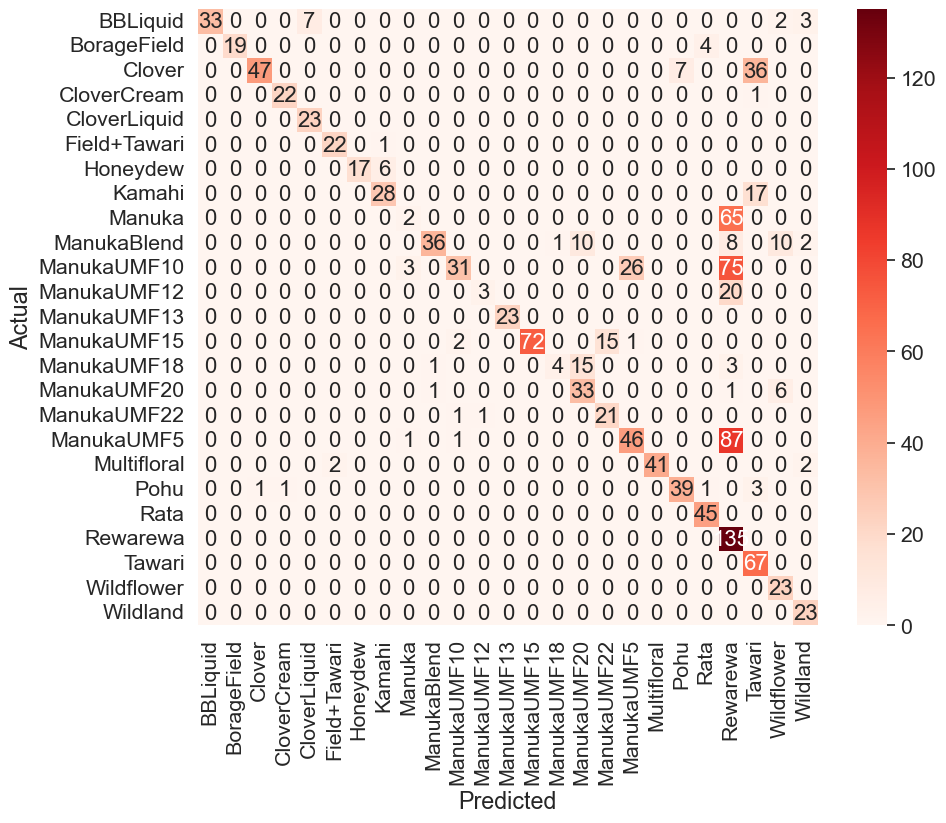}
		\caption{Confusion matrix of Replay on Task 4.}
		\label{replay44}
	\end{figure}
	
	\begin{figure}[H]
		\centering
		\includegraphics[scale=0.7]{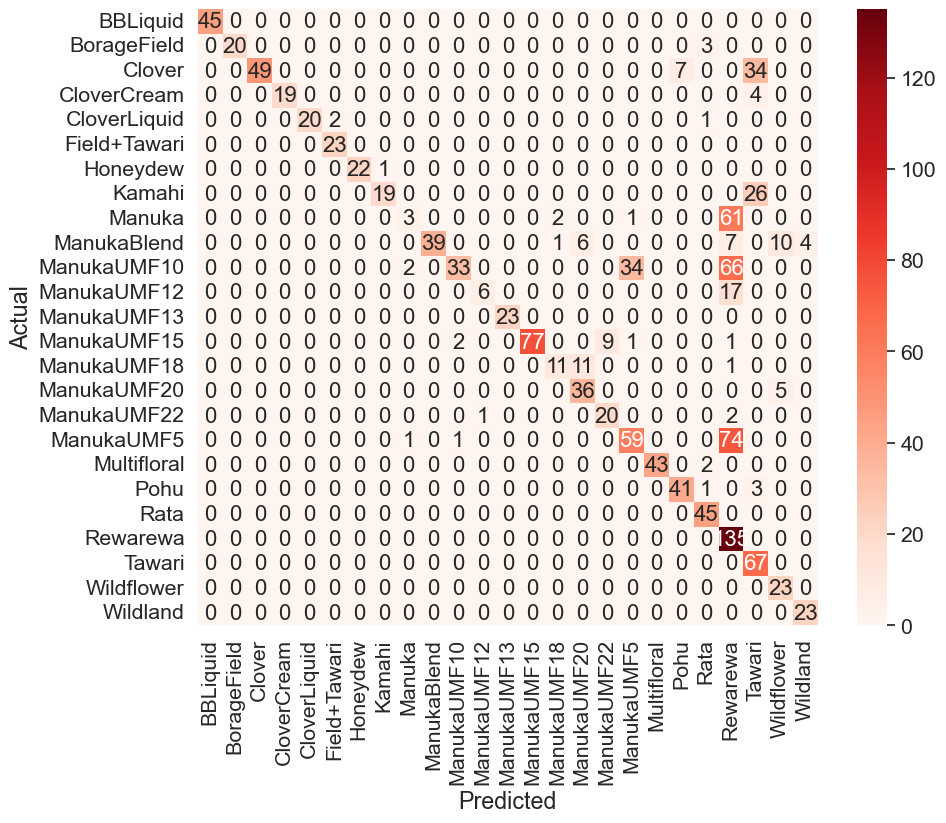}
		\caption{Confusion matrix of iCarl on Task 4.}
		\label{icarl44}
	\end{figure}
	
	\begin{figure}[H]
		\centering
		\includegraphics[scale=0.7]{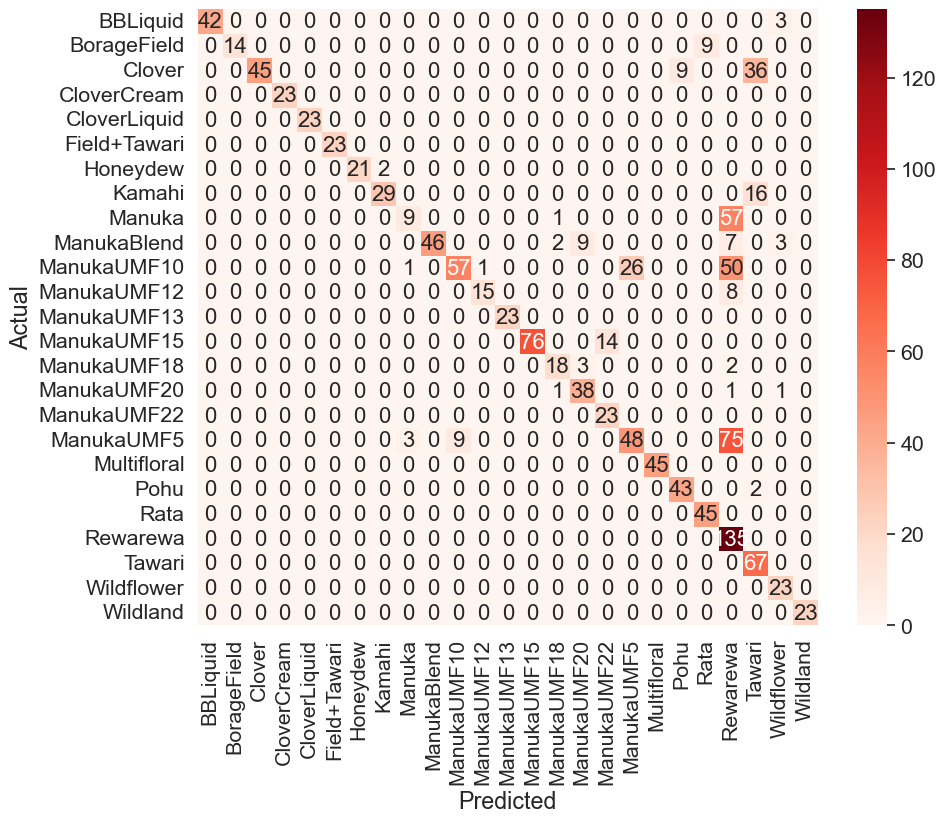}
		\caption{Confusion matrix of WA on Task 4.}
		\label{wa44}
	\end{figure}

	\newpage
	
	\section{\textbf{\Large Sensitivity Analysis of CB's hyperparameters: Replacement Rate and Maturity Threshold}}\label{cilsensitive}

	\begin{table}[H]
		\centering
		\begin{tabular}{|c|c|c|c|c|c|}
			\hline
			\textbf{CB}&\multicolumn{5}{c|}{\textbf{Weighted Average F1 Score of Finetune + CB in Incremental Task No.}} \\
			\hline
			Parameters & 0 & 1 & 2 & 3 & 4 \\
			\hline
			0.001 / 500 & 100.00\%$\pm$0.00\% & 48.39\%$\pm$7.86\% & 24.29\%$\pm$3.71\% & 13.10\%$\pm$0.10\% & 8.58\%$\pm$0.15\% \\ \hline
			0.001 / 2000 & 100.00\%$\pm$0.00\% & 44.19\%$\pm$9.28\% & 24.94\%$\pm$2.33\% & 13.97\%$\pm$1.39\% & 9.28\%$\pm$0.50\% \\ \hline
			0.001 / 5000 & 99.90\%$\pm$0.22\% & 39.04\%$\pm$5.89\% & 25.00\%$\pm$3.17\% & 14.20\%$\pm$1.55\% & 10.70\%$\pm$0.92\% \\ \hline
			0.1 / 500 & 100.00\%$\pm$0.00\% & 39.35\%$\pm$4.41\% & 25.29\%$\pm$3.32\% & 12.90\%$\pm$0.07\% & 8.71\%$\pm$0.18\% \\ \hline
			0.1 / 2000 & 100.00\%$\pm$0.00\% & 43.03\%$\pm$7.12\% & 26.42\%$\pm$2.83\% & 13.63\%$\pm$0.86\% & 11.72\%$\pm$2.62\% \\ \hline
			0.1 / 5000 & 100.00\%$\pm$0.00\% & 42.06\%$\pm$4.33\% & 26.70\%$\pm$2.10\% & 14.41\%$\pm$1.78\% & 11.37\%$\pm$1.82\% \\ \hline
			0.5 / 500 & 100.00\%$\pm$0.00\% & 45.03\%$\pm$7.35\% & 25.61\%$\pm$2.81\% & 12.96\%$\pm$0.07\% & 8.53\%$\pm$0.13\% \\ \hline
			0.5 / 2000 & 100.00\%$\pm$0.00\% & 40.50\%$\pm$6.43\% & 22.29\%$\pm$4.30\% & 14.11\%$\pm$1.51\% & 9.83\%$\pm$1.76\% \\ \hline
			0.5 / 5000 & 100.00\%$\pm$0.00\% & 42.54\%$\pm$5.82\% & 26.19\%$\pm$3.82\% & 13.58\%$\pm$1.44\% & 11.36\%$\pm$0.66\% \\ \hline 
		\end{tabular}
		\caption{Hyperparameter Sensitive Analysis of CB with Fintune. The first column shows the different combinations of Replacement Rate and Maturity Threshold.}
		\label{hpfinetune}
	\end{table}
	
	\begin{table}[H]
		\centering
		\begin{tabular}{|c|c|c|c|c|c|}
			\hline
			\textbf{CB}&\multicolumn{5}{c|}{\textbf{Weighted Average F1 Score of LwF + CB in Incremental Task No.}} \\
			\hline
			Parameters & 0 & 1 & 2 & 3 & 4 \\
			\hline
			0.001 / 500 & 100.00\%$\pm$0.00\% & 42.28\%$\pm$8.70\% & 24.99\%$\pm$3.95\% & 13.02\%$\pm$0.17\% & 8.54\%$\pm$0.15\% \\ \hline
			0.001 / 2000 & 100.00\%$\pm$0.00\% & 47.47\%$\pm$9.26\% & 24.26\%$\pm$6.93\% & 15.16\%$\pm$2.13\% & 12.58\%$\pm$0.80\% \\ \hline
			0.001 / 5000 & 100.00\%$\pm$0.00\% & 43.25\%$\pm$6.92\% & 23.49\%$\pm$4.33\% & 14.58\%$\pm$2.30\% & 13.40\%$\pm$1.53\% \\ \hline
			0.1 / 500 & 100.00\%$\pm$0.00\% & 49.59\%$\pm$6.10\% & 23.36\%$\pm$2.63\% & 12.89\%$\pm$0.02\% & 8.62\%$\pm$0.06\% \\ \hline
			0.1 / 2000 & 100.00\%$\pm$0.00\% & 41.20\%$\pm$6.21\% & 23.72\%$\pm$1.70\% & 15.45\%$\pm$1.43\% & 13.06\%$\pm$1.77\% \\ \hline
			0.1 / 5000 & 100.00\%$\pm$0.00\% & 44.06\%$\pm$6.89\% & 24.61\%$\pm$3.64\% & 14.39\%$\pm$1.39\% & 11.30\%$\pm$0.98\% \\ \hline
			0.5 / 500 & 100.00\%$\pm$0.00\% & 42.46\%$\pm$4.97\% & 22.19\%$\pm$4.51\% & 12.87\%$\pm$0.03\% & 8.50\%$\pm$0.11\% \\ \hline
			0.5 / 2000 & 100.00\%$\pm$0.00\% & 40.38\%$\pm$4.99\% & 25.25\%$\pm$1.09\% & 15.49\%$\pm$2.25\% & 11.23\%$\pm$1.64\% \\ \hline
			0.5 / 5000 & 100.00\%$\pm$0.00\% & 39.24\%$\pm$4.01\% & 22.25\%$\pm$2.86\% & 14.94\%$\pm$1.39\% & 11.27\%$\pm$1.63\% \\ \hline
		\end{tabular}
		\caption{Hyperparameter Sensitive Analysis of CB with LwF. The first column shows the different combinations of Replacement Rate and Maturity Threshold.}
		\label{hplwf}
	\end{table}
	
	\begin{table}[H]
		\centering
		\begin{tabular}{|c|c|c|c|c|c|}
			\hline
			\textbf{CB}&\multicolumn{5}{c|}{\textbf{Weighted Average F1 Score of EWC + CB in Incremental Task No.}} \\
			\hline
			Parameters & 0 & 1 & 2 & 3 & 4 \\
			\hline
			0.001 / 500 & 100.00\%$\pm$0.00\% & 42.37\%$\pm$11.13\% & 26.09\%$\pm$4.13\% & 13.04\%$\pm$0.10\% & 8.49\%$\pm$0.10\% \\ \hline
			0.001 / 2000 & 100.00\%$\pm$0.00\% & 41.81\%$\pm$6.36\% & 23.99\%$\pm$3.97\% & 14.01\%$\pm$1.03\% & 10.90\%$\pm$2.67\% \\ \hline
			0.001 / 5000 & 100.00\%$\pm$0.00\% & 46.53\%$\pm$9.01\% & 22.31\%$\pm$3.15\% & 15.73\%$\pm$2.62\% & 9.84\%$\pm$0.60\% \\ \hline
			0.1 / 500 & 100.00\%$\pm$0.00\% & 45.46\%$\pm$6.18\% & 26.03\%$\pm$2.43\% & 12.91\%$\pm$0.07\% & 8.64\%$\pm$0.12\% \\ \hline
			0.1 / 2000 & 100.00\%$\pm$0.00\% & 45.42\%$\pm$4.54\% & 24.20\%$\pm$1.15\% & 13.92\%$\pm$1.31\% & 9.64\%$\pm$1.38\% \\ \hline
			0.1 / 5000 & 100.00\%$\pm$0.00\% & 44.16\%$\pm$5.12\% & 26.41\%$\pm$2.12\% & 14.91\%$\pm$2.25\% & 9.18\%$\pm$0.97\% \\ \hline
			0.5 / 500 & 99.90\%$\pm$0.22\% & 49.23\%$\pm$4.93\% & 25.20\%$\pm$3.34\% & 12.95\%$\pm$0.12\% & 8.42\%$\pm$0.08\% \\ \hline
			0.5 / 2000 & 100.00\%$\pm$0.00\% & 39.37\%$\pm$4.43\% & 26.64\%$\pm$3.22\% & 14.46\%$\pm$1.99\% & 9.86\%$\pm$1.38\% \\ \hline
			0.5 / 5000 & 100.00\%$\pm$0.00\% & 41.69\%$\pm$4.18\% & 21.42\%$\pm$1.78\% & 14.04\%$\pm$1.97\% & 10.65\%$\pm$1.57\% \\ \hline
			
		\end{tabular}
		\caption{Hyperparameter Sensitive Analysis of CB with EWC. The first column shows the different combinations of Replacement Rate and Maturity Threshold.}
		\label{hpewc}
	\end{table}
	
	\begin{table}[H]
		\centering
		\begin{tabular}{|c|c|c|c|c|c|}
			\hline
			\textbf{CB}&\multicolumn{5}{c|}{\textbf{Weighted Average F1 Score of Replay + CB in Incremental Task No.}} \\
			\hline
			Parameters & 0 & 1 & 2 & 3 & 4 \\
			\hline
			0.001 / 500 & 100.00\%$\pm$0.00\% & 81.07\%$\pm$2.94\% & 57.54\%$\pm$3.83\% & 62.39\%$\pm$1.99\% & 65.14\%$\pm$6.26\% \\ \hline
			0.001 / 2000 & 100.00\%$\pm$0.00\% & 79.55\%$\pm$6.71\% & 59.95\%$\pm$7.25\% & 68.51\%$\pm$2.73\% & 65.53\%$\pm$3.94\% \\ \hline
			0.001 / 5000 & 100.00\%$\pm$0.00\% & 81.03\%$\pm$3.93\% & 63.29\%$\pm$7.46\% & 65.93\%$\pm$5.64\% & 65.66\%$\pm$5.12\% \\ \hline
			0.1 / 500 & 100.00\%$\pm$0.00\% & 83.15\%$\pm$1.89\% & 52.73\%$\pm$6.91\% & 60.17\%$\pm$13.30\% & 63.89\%$\pm$3.61\% \\ \hline
			0.1 / 2000 & 100.00\%$\pm$0.00\% & 82.31\%$\pm$3.53\% & 57.37\%$\pm$8.21\% & 68.01\%$\pm$3.36\% & 67.95\%$\pm$4.08\% \\ \hline
			0.1 / 5000 & 100.00\%$\pm$0.00\% & 80.65\%$\pm$5.24\% & 56.74\%$\pm$6.25\% & 62.90\%$\pm$8.52\% & 63.13\%$\pm$9.94\% \\ \hline
			0.5 / 500 & 100.00\%$\pm$0.00\% & 79.59\%$\pm$6.63\% & 47.02\%$\pm$11.39\% & 63.60\%$\pm$7.64\% & 66.91\%$\pm$4.17\% \\ \hline
			0.5 / 2000 & 100.00\%$\pm$0.00\% & 83.86\%$\pm$4.81\% & 66.18\%$\pm$6.03\% & 73.20\%$\pm$2.04\% & 70.11\%$\pm$2.13\% \\ \hline
			0.5 / 5000 & 100.00\%$\pm$0.00\% & 76.67\%$\pm$7.27\% & 55.15\%$\pm$12.23\% & 66.52\%$\pm$4.83\% & 65.80\%$\pm$2.77\% \\ \hline
			
		\end{tabular}
		\caption{Hyperparameter Sensitive Analysis of CB with Replay. The first column shows the different combinations of Replacement Rate and Maturity Threshold.}
		\label{hpreplay}
	\end{table}

	\begin{table}[H]
		\centering
		\begin{tabular}{|c|c|c|c|c|c|}
			\hline
			\textbf{CB}&\multicolumn{5}{c|}{\textbf{Weighted Average F1 Score of iCarl + CB in Incremental Task No.}} \\
			\hline
			Parameters & 0 & 1 & 2 & 3 & 4 \\
			\hline
			0.001 / 500 & 100.00\%$\pm$0.00\% & 82.01\%$\pm$2.46\% & 56.50\%$\pm$5.16\% & 58.59\%$\pm$5.05\% & 66.01\%$\pm$6.33\% \\ \hline
			0.001 / 2000 & 100.00\%$\pm$0.00\% & 78.45\%$\pm$7.22\% & 57.77\%$\pm$8.47\% & 61.62\%$\pm$8.05\% & 61.51\%$\pm$8.63\% \\ \hline
			0.001 / 5000 & 100.00\%$\pm$0.00\% & 83.24\%$\pm$1.67\% & 55.35\%$\pm$3.86\% & 68.13\%$\pm$1.98\% & 71.89\%$\pm$1.68\% \\ \hline
			0.1 / 500 & 99.90\%$\pm$0.22\% & 77.65\%$\pm$6.21\% & 47.00\%$\pm$4.57\% & 68.75\%$\pm$4.16\% & 63.95\%$\pm$5.21\% \\ \hline
			0.1 / 2000 & 100.00\%$\pm$0.00\% & 78.14\%$\pm$2.18\% & 53.24\%$\pm$10.91\% & 65.39\%$\pm$2.00\% & 68.03\%$\pm$3.50\% \\ \hline
			0.1 / 5000 & 100.00\%$\pm$0.00\% & 79.71\%$\pm$2.55\% & 57.96\%$\pm$4.06\% & 66.79\%$\pm$1.96\% & 67.41\%$\pm$4.75\% \\ \hline
			0.5 / 500 & 100.00\%$\pm$0.00\% & 83.78\%$\pm$4.98\% & 49.94\%$\pm$9.39\% & 61.60\%$\pm$4.68\% & 63.05\%$\pm$5.15\% \\ \hline
			0.5 / 2000 & 100.00\%$\pm$0.00\% & 81.66\%$\pm$5.03\% & 59.33\%$\pm$7.74\% & 65.64\%$\pm$4.05\% & 67.81\%$\pm$5.37\% \\ \hline
			0.5 / 5000 & 100.00\%$\pm$0.00\% & 76.75\%$\pm$4.98\% & 55.37\%$\pm$6.60\% & 64.44\%$\pm$8.43\% & 64.18\%$\pm$8.11\% \\ \hline 
			
		\end{tabular}
		\caption{Hyperparameter Sensitive Analysis of CB with iCarl. The first column shows the different combinations of Replacement Rate and Maturity Threshold.}
		\label{hpicarl}
	\end{table}
	
	\begin{table}[H]
		\centering
		\begin{tabular}{|c|c|c|c|c|c|}
			\hline
			\textbf{CB}&\multicolumn{5}{c|}{\textbf{Weighted Average F1 Score of WA + CB in Incremental Task No.}} \\
			\hline
			& 0 & 1 & 2 & 3 & 4 \\
			\hline
			0.001 / 500 & 100.00\%$\pm$0.00\% & 80.16\%$\pm$3.17\% & 50.85\%$\pm$5.46\% & 59.65\%$\pm$3.65\% & 67.88\%$\pm$3.90\% \\ \hline
			0.001 / 2000 & 100.00\%$\pm$0.00\% & 81.58\%$\pm$3.19\% & 55.53\%$\pm$7.75\% & 69.00\%$\pm$2.83\% & 68.84\%$\pm$3.42\% \\ \hline
			0.001 / 5000 & 100.00\%$\pm$0.00\% & 80.24\%$\pm$5.38\% & 54.24\%$\pm$11.15\% & 66.59\%$\pm$6.44\% & 67.69\%$\pm$5.08\% \\ \hline
			0.1 / 500 & 100.00\%$\pm$0.00\% & 82.61\%$\pm$7.34\% & 50.35\%$\pm$7.22\% & 64.82\%$\pm$4.81\% & 62.43\%$\pm$3.26\% \\ \hline
			0.1 / 2000 & 100.00\%$\pm$0.00\% & 81.58\%$\pm$5.73\% & 58.43\%$\pm$7.46\% & 67.01\%$\pm$6.22\% & 67.80\%$\pm$5.85\% \\ \hline
			0.1 / 5000 & 100.00\%$\pm$0.00\% & 82.86\%$\pm$7.80\% & 59.43\%$\pm$5.38\% & 66.47\%$\pm$6.54\% & 65.83\%$\pm$8.67\% \\ \hline
			0.5 / 500 & 100.00\%$\pm$0.00\% & 81.83\%$\pm$4.32\% & 49.67\%$\pm$5.19\% & 63.89\%$\pm$4.83\% & 63.70\%$\pm$2.80\% \\ \hline
			0.5 / 2000 & 100.00\%$\pm$0.00\% & 79.23\%$\pm$3.33\% & 51.97\%$\pm$4.88\% & 62.83\%$\pm$5.17\% & 63.55\%$\pm$3.41\% \\ \hline
			0.5 / 5000 & 100.00\%$\pm$0.00\% & 84.07\%$\pm$1.37\% & 58.09\%$\pm$7.52\% & 64.02\%$\pm$5.09\% & 68.06\%$\pm$4.12\% \\ \hline
			
		\end{tabular}
		\caption{Hyperparameter Sensitive Analysis of CB with WA. The first column shows the different combinations of Replacement Rate and Maturity Threshold.}
		\label{hpwa}
	\end{table}
	
	\begin{table}[H]
		\centering
		\begin{tabular}{|c|c|c|c|c|c|}
			\hline
			\textbf{CB}&\multicolumn{5}{c|}{\textbf{Weighted Average F1 Score of DER + CB in Incremental Task No.}} \\
			\hline
			Parameters & 0 & 1 & 2 & 3 & 4 \\
			\hline
			0.001 / 500 & 100.00\%$\pm$0.00\% & 88.23\%$\pm$4.52\% & 66.91\%$\pm$5.80\% & 70.99\%$\pm$1.84\% & 69.22\%$\pm$3.14\% \\ \hline
			0.001 / 2000 & 100.00\%$\pm$0.00\% & 85.60\%$\pm$5.46\% & 67.72\%$\pm$5.55\% & 70.08\%$\pm$3.13\% & 70.60\%$\pm$4.56\% \\ \hline
			0.001 / 5000 & 100.00\%$\pm$0.00\% & 87.98\%$\pm$2.24\% & 71.65\%$\pm$4.87\% & 69.77\%$\pm$3.80\% & 71.38\%$\pm$2.84\% \\ \hline
			0.1 / 500 & 100.00\%$\pm$0.00\% & 86.98\%$\pm$0.72\% & 69.31\%$\pm$2.39\% & 71.51\%$\pm$2.47\% & 69.72\%$\pm$1.68\% \\ \hline
			0.1 / 2000 & 100.00\%$\pm$0.00\% & 86.64\%$\pm$2.90\% & 72.50\%$\pm$1.85\% & 73.13\%$\pm$2.08\% & 75.73\%$\pm$5.48\% \\ \hline
			0.1 / 5000 & 100.00\%$\pm$0.00\% & 86.81\%$\pm$1.36\% & 69.18\%$\pm$5.07\% & 69.58\%$\pm$2.21\% & 71.09\%$\pm$3.37\% \\ \hline
			0.5 / 500 & 100.00\%$\pm$0.00\% & 83.88\%$\pm$2.86\% & 72.89\%$\pm$5.89\% & 69.40\%$\pm$2.81\% & 72.24\%$\pm$4.51\% \\ \hline
			0.5 / 2000 & 100.00\%$\pm$0.00\% & 87.94\%$\pm$4.12\% & 67.19\%$\pm$7.62\% & 69.25\%$\pm$3.65\% & 71.43\%$\pm$1.98\% \\ \hline
			0.5 / 5000 & 100.00\%$\pm$0.00\% & 88.45\%$\pm$2.45\% & 69.17\%$\pm$3.48\% & 71.51\%$\pm$3.63\% & 72.72\%$\pm$2.58\% \\ \hline 
			
		\end{tabular}
		\caption{Hyperparameter Sensitive Analysis of CB with DER. The first column shows the different combinations of Replacement Rate and Maturity Threshold.}
		\label{hpder}
	\end{table}
	
	\begin{table}[H]
		\centering
		\begin{tabular}{|c|c|c|c|c|c|}
			\hline
			\textbf{CB}&\multicolumn{5}{c|}{\textbf{Weighted Average F1 Score of FOSTER + CB in Incremental Task No.}} \\
			\hline
			Parameters & 0 & 1 & 2 & 3 & 4 \\
			\hline
			0.001 / 500 & 99.90\%$\pm$0.22\% & 94.03\%$\pm$2.01\% & 57.53\%$\pm$8.67\% & 72.83\%$\pm$6.87\% & 69.02\%$\pm$5.52\% \\ \hline
			0.001 / 2000 & 100.00\%$\pm$0.00\% & 94.91\%$\pm$1.64\% & 64.09\%$\pm$9.91\% & 77.89\%$\pm$2.58\% & 71.94\%$\pm$3.19\% \\ \hline
			0.001 / 5000 & 100.00\%$\pm$0.00\% & 96.14\%$\pm$1.72\% & 59.33\%$\pm$6.76\% & 77.49\%$\pm$2.81\% & 70.52\%$\pm$2.24\% \\ \hline
			0.1 / 500 & 100.00\%$\pm$0.00\% & 97.23\%$\pm$1.61\% & 67.33\%$\pm$6.15\% & 79.88\%$\pm$2.41\% & 73.28\%$\pm$2.55\% \\ \hline
			0.1 / 2000 & 100.00\%$\pm$0.00\% & 95.03\%$\pm$1.82\% & 62.75\%$\pm$5.65\% & 74.51\%$\pm$2.38\% & 67.75\%$\pm$4.60\% \\ \hline
			0.1 / 5000 & 100.00\%$\pm$0.00\% & 96.88\%$\pm$0.79\% & 64.27\%$\pm$3.99\% & 78.91\%$\pm$3.31\% & 71.28\%$\pm$2.25\% \\ \hline
			0.5 / 500 & 100.00\%$\pm$0.00\% & 95.60\%$\pm$1.82\% & 58.93\%$\pm$5.92\% & 75.70\%$\pm$5.12\% & 72.15\%$\pm$3.67\% \\ \hline
			0.5 / 2000 & 100.00\%$\pm$0.00\% & 97.20\%$\pm$0.99\% & 68.87\%$\pm$6.33\% & 79.20\%$\pm$3.96\% & 74.34\%$\pm$3.51\% \\ \hline
			0.5 / 5000 & 100.00\%$\pm$0.00\% & 95.81\%$\pm$1.70\% & 64.88\%$\pm$9.71\% & 76.54\%$\pm$3.10\% & 70.64\%$\pm$3.76\% \\ \hline
			
		\end{tabular}
		\caption{Hyperparameter Sensitive Analysis of CB with FOSTER. The first column shows the different combinations of Replacement Rate and Maturity Threshold.}
		\label{hpfoster}
	\end{table}
	
	\begin{table}[H]
		\centering
		\begin{tabular}{|c|c|c|c|c|c|}
			\hline
			\textbf{CB}&\multicolumn{5}{c|}{\textbf{Weighted Average F1 Score of MEMO + CB in Incremental Task No.}} \\
			\hline
			Parameters & 0 & 1 & 2 & 3 & 4 \\
			\hline
			0.001 500 & 100.00\%$\pm$0.00\% & 81.17\%$\pm$3.11\% & 54.56\%$\pm$5.58\% & 67.40\%$\pm$8.14\% & 67.67\%$\pm$9.23\% \\ \hline
			0.001 2000 & 100.00\%$\pm$0.00\% & 82.31\%$\pm$2.94\% & 60.42\%$\pm$5.41\% & 71.54\%$\pm$1.51\% & 70.94\%$\pm$3.13\% \\ \hline
			0.001 5000 & 100.00\%$\pm$0.00\% & 84.18\%$\pm$2.37\% & 59.79\%$\pm$9.58\% & 69.15\%$\pm$3.28\% & 69.96\%$\pm$4.21\% \\ \hline
			0.1 500 & 100.00\%$\pm$0.00\% & 77.97\%$\pm$2.91\% & 46.85\%$\pm$5.82\% & 61.11\%$\pm$6.78\% & 70.20\%$\pm$2.83\% \\ \hline
			0.1 2000 & 100.00\%$\pm$0.00\% & 77.04\%$\pm$3.35\% & 54.57\%$\pm$10.21\% & 68.46\%$\pm$2.86\% & 69.69\%$\pm$1.93\% \\ \hline
			0.1 5000 & 100.00\%$\pm$0.00\% & 81.44\%$\pm$2.97\% & 61.32\%$\pm$5.95\% & 67.40\%$\pm$3.39\% & 67.89\%$\pm$3.53\% \\ \hline
			0.5 500 & 100.00\%$\pm$0.00\% & 81.27\%$\pm$2.34\% & 52.56\%$\pm$5.13\% & 67.74\%$\pm$7.21\% & 66.80\%$\pm$1.64\% \\ \hline
			0.5 2000 & 100.00\%$\pm$0.00\% & 78.10\%$\pm$1.46\% & 57.59\%$\pm$6.72\% & 63.02\%$\pm$3.02\% & 63.18\%$\pm$2.55\% \\ \hline
			0.5 5000 & 100.00\%$\pm$0.00\% & 80.54\%$\pm$2.42\% & 58.99\%$\pm$4.74\% & 68.19\%$\pm$3.41\% & 67.08\%$\pm$1.76\% \\ \hline 
			
		\end{tabular}
		\caption{Hyperparameter Sensitive Analysis of CB with MEMO. The first column shows the different combinations of Replacement Rate and Maturity Threshold.}
		\label{hpmemo}
	\end{table}
	
	\newpage
	
	\section{\textbf{\Large Kernel Density Estimation of CIL Methods}}\label{cilkde}
	
	\begin{figure}[H]
		\hspace{-0.3in}
		\includegraphics[scale=0.75]{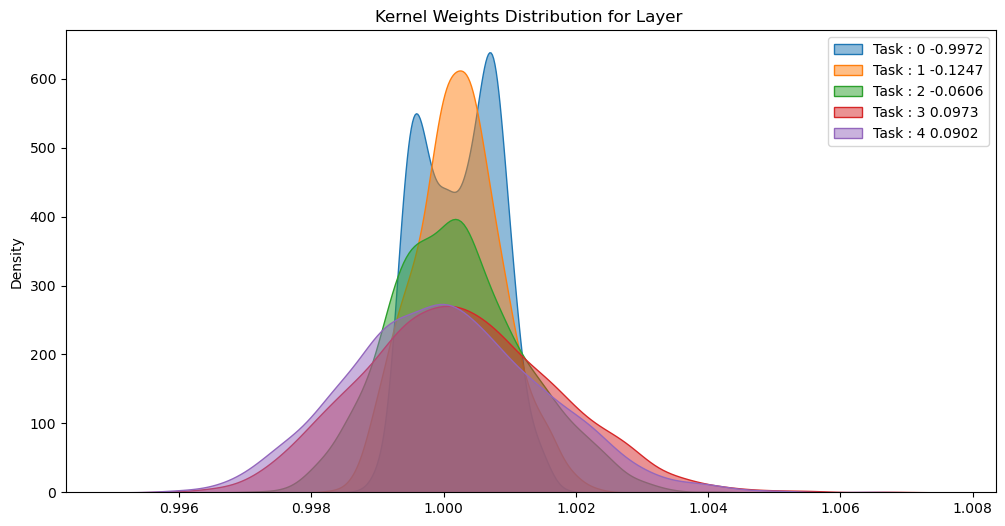}
		\caption{Kernel Density Estimation of Finetune}
		\label{kft}
	\end{figure}
	
	\begin{figure}[H]
		\hspace{-0.3in}
		\includegraphics[scale=0.75]{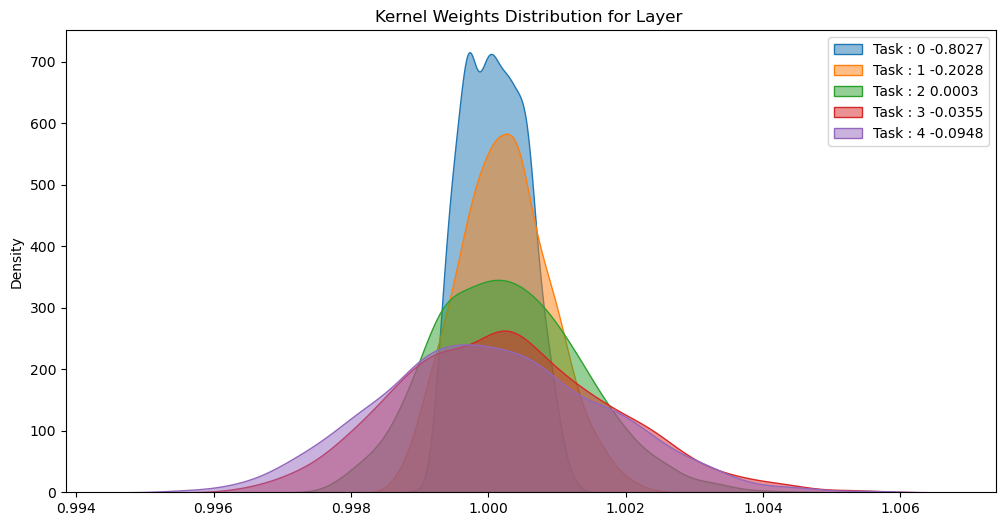}
		\caption{Kernel Density Estimation of Finetune + CB}
		\label{kftcb}
	\end{figure}
	
	\begin{figure}[H]
		\hspace{-0.3in}
		\includegraphics[scale=0.75]{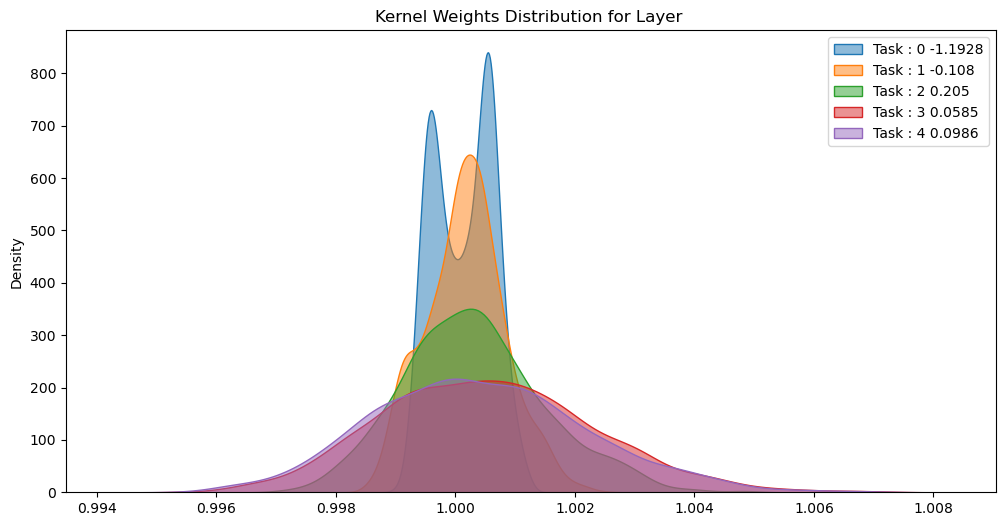}
		\caption{Kernel Density Estimation of LwF}
		\label{klwf}
	\end{figure}
	
	\begin{figure}[H]
		\hspace{-0.3in}
		\includegraphics[scale=0.75]{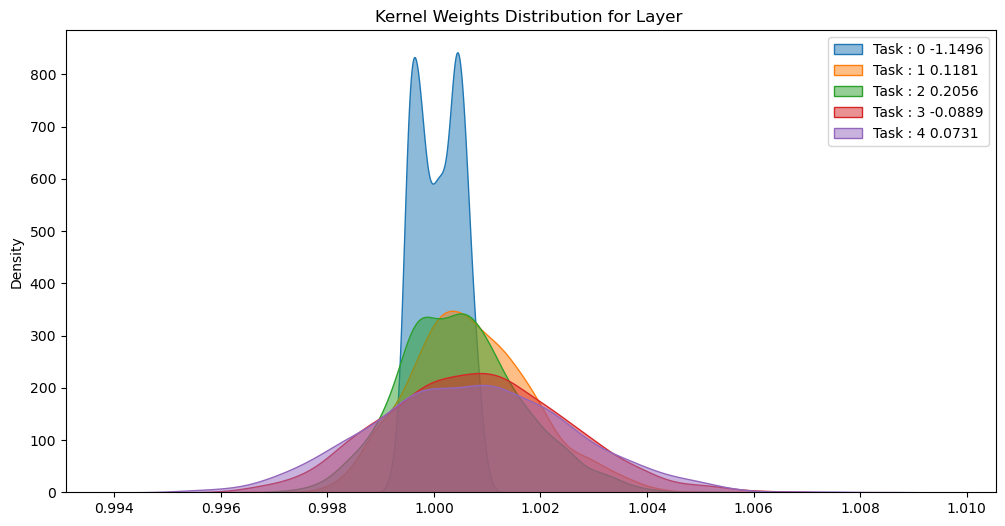}
		\caption{Kernel Density Estimation of LwF + CB}
		\label{klwfcb}
	\end{figure}
	
	\begin{figure}[H]
		\hspace{-0.3in}
		\includegraphics[scale=0.75]{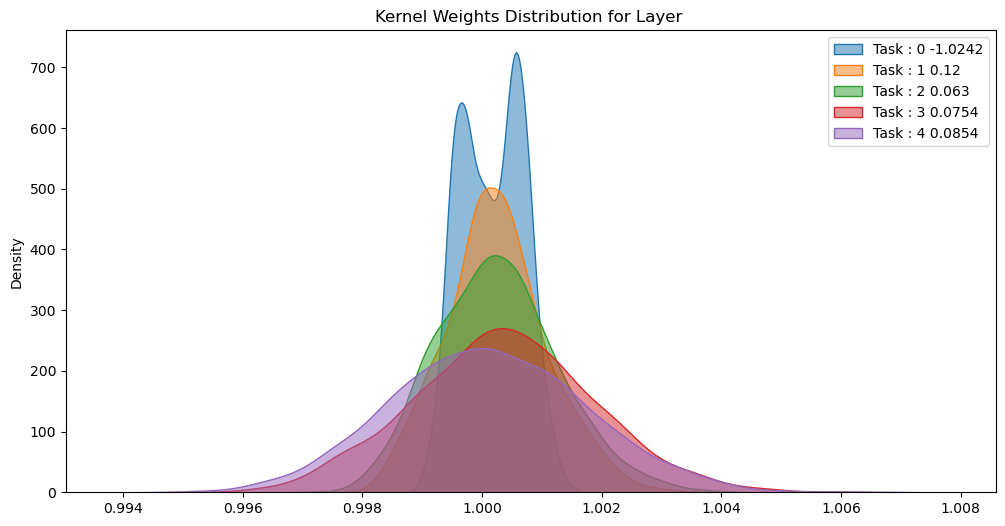}
		\caption{Kernel Density Estimation of EWC}
		\label{kewc}
	\end{figure}
	
	\begin{figure}[H]
		\hspace{-0.3in}
		\includegraphics[scale=0.75]{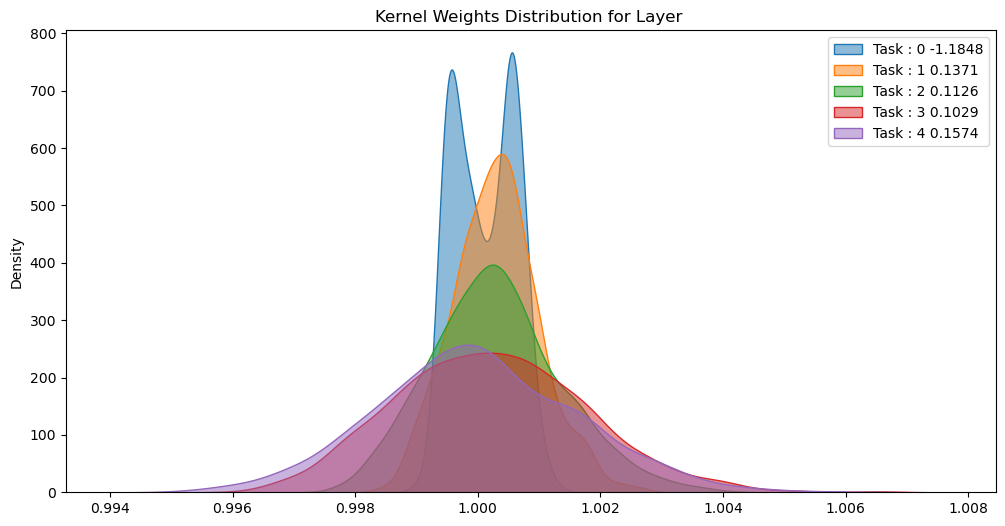}
		\caption{Kernel Density Estimation of EWC + CB}
		\label{kewccb}
	\end{figure}
	
	\begin{figure}[H]
		\hspace{-0.3in}
		\includegraphics[scale=0.75]{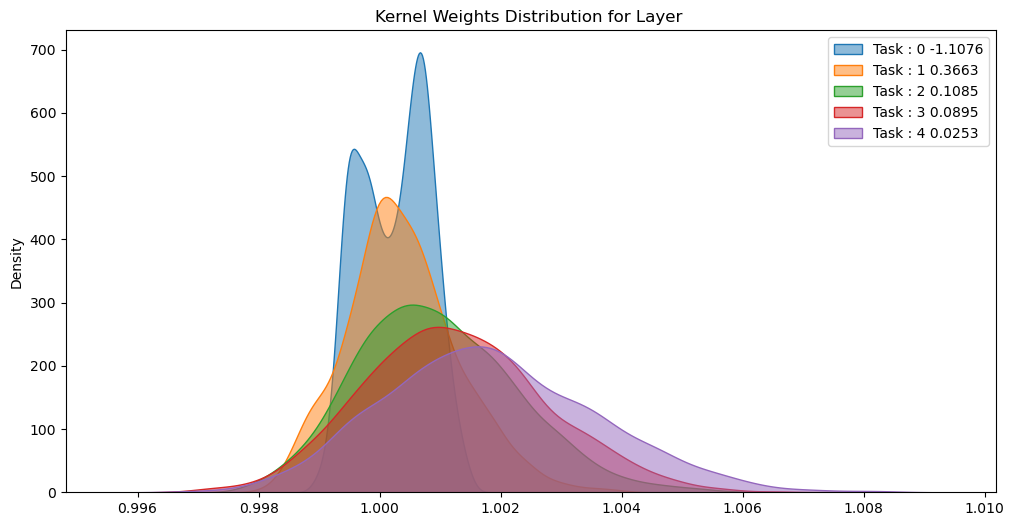}
		\caption{Kernel Density Estimation of Replay}
		\label{kreplay}
	\end{figure}
	
	\begin{figure}[H]
		\hspace{-0.3in}
		\includegraphics[scale=0.75]{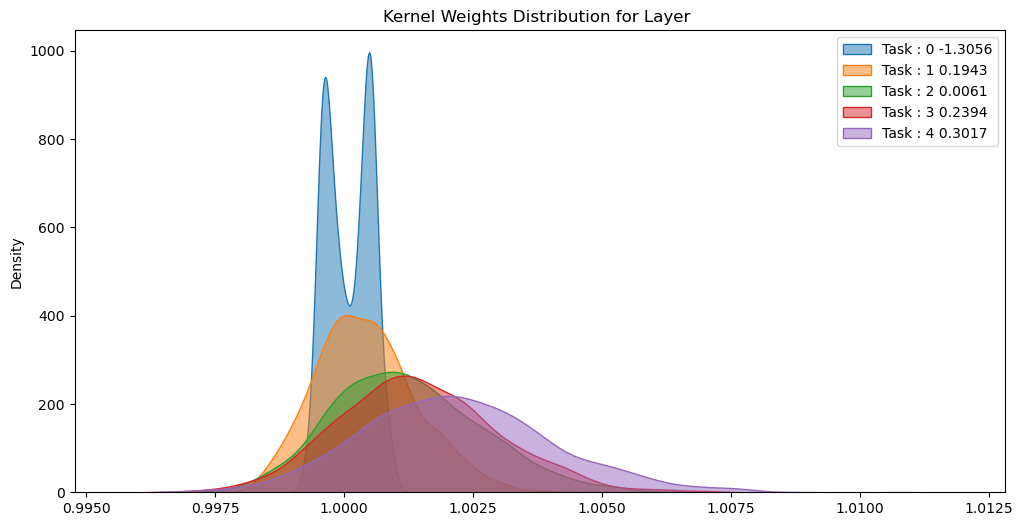}
		\caption{Kernel Density Estimation of Replay + CB}
		\label{kreplaycb}
	\end{figure}

	\begin{figure}[H]
		\hspace{-0.3in}
		\includegraphics[scale=0.75]{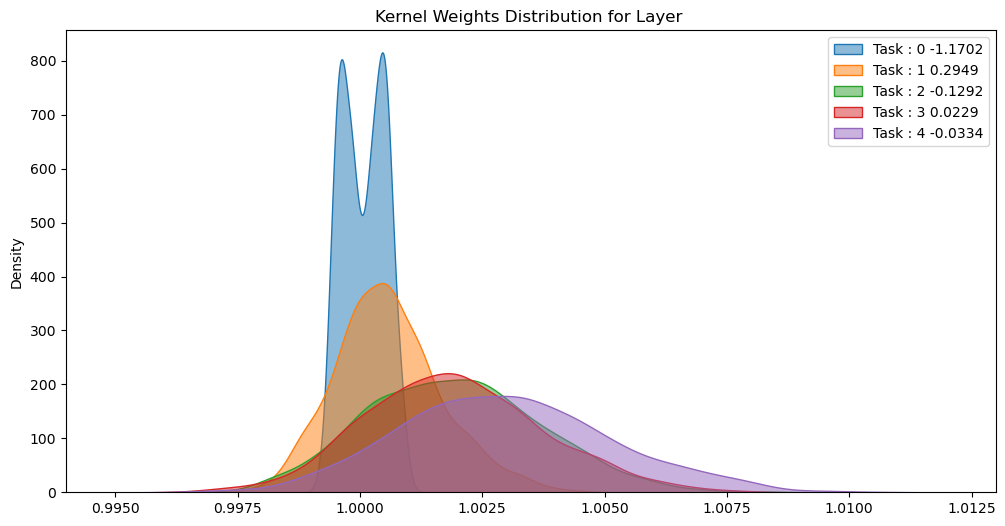}
		\caption{Kernel Density Estimation of iCarl}
		\label{kicarl}
	\end{figure}
	
	\begin{figure}[H]
		\hspace{-0.3in}
		\includegraphics[scale=0.75]{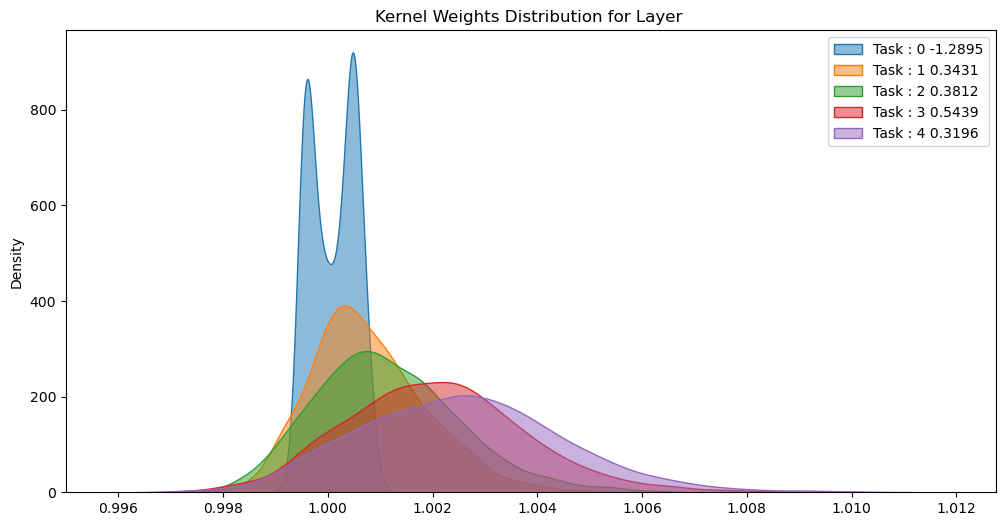}
		\caption{Kernel Density Estimation of iCarl + CB}
		\label{kicarlcb}
	\end{figure}
	
	\begin{figure}[H]
		\hspace{-0.3in}
		\includegraphics[scale=0.75]{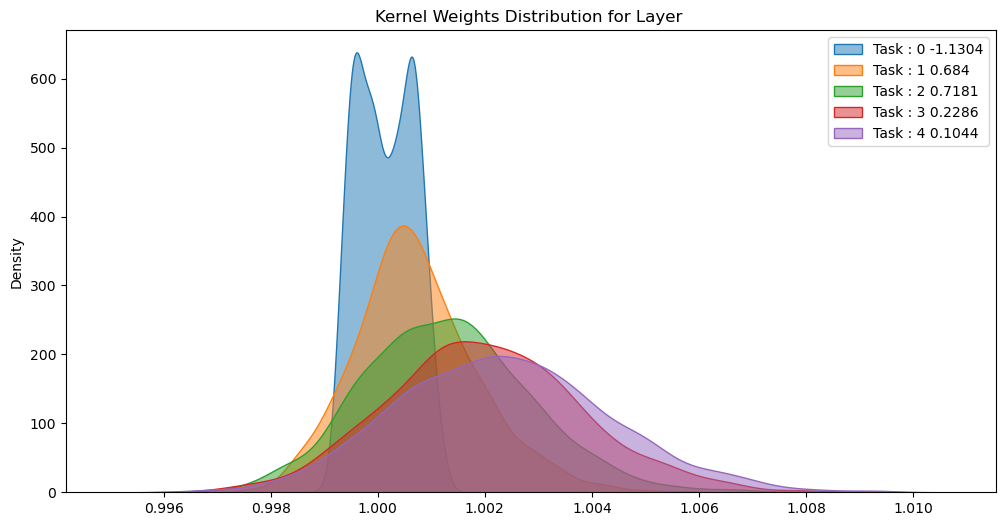}
		\caption{Kernel Density Estimation of WA}
		\label{kwa}
	\end{figure}
	
	\begin{figure}[H]
		\hspace{-0.3in}
		\includegraphics[scale=0.75]{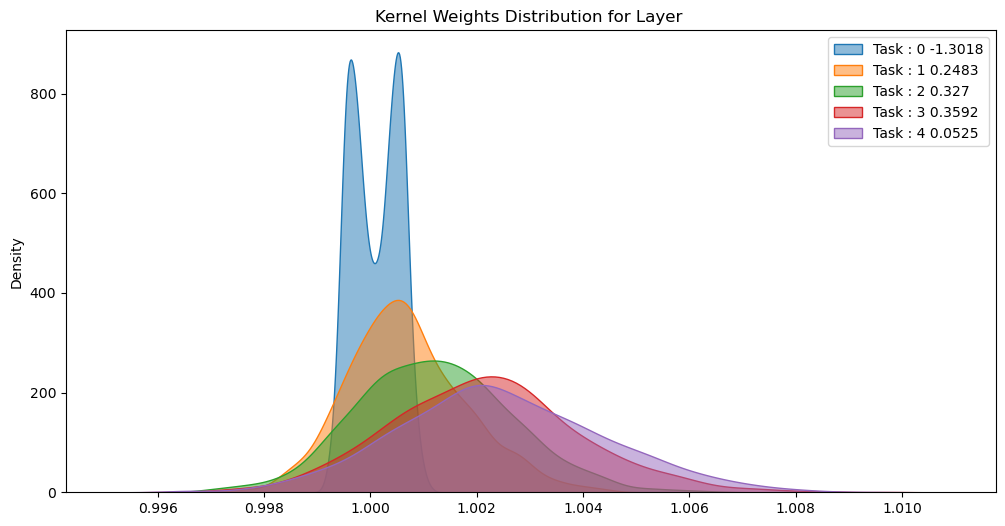}
		\caption{Kernel Density Estimation of WA + CB}
		\label{kwacb}
	\end{figure}
	
	\begin{figure}[H]
		\hspace{-0.3in}
		\includegraphics[scale=0.75]{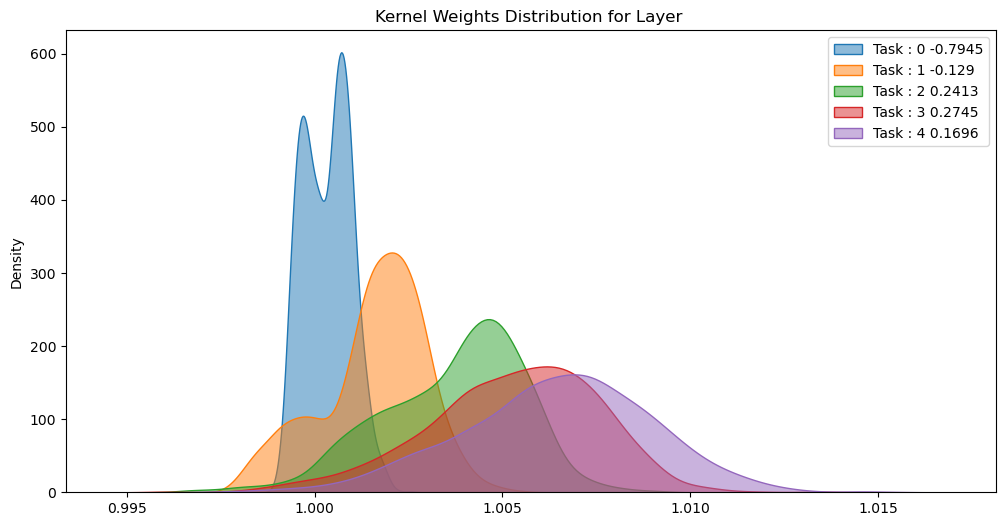}
		\caption{Kernel Density Estimation of DER}
		\label{kder}
	\end{figure}
	
	\begin{figure}[H]
		\hspace{-0.3in}
		\includegraphics[scale=0.75]{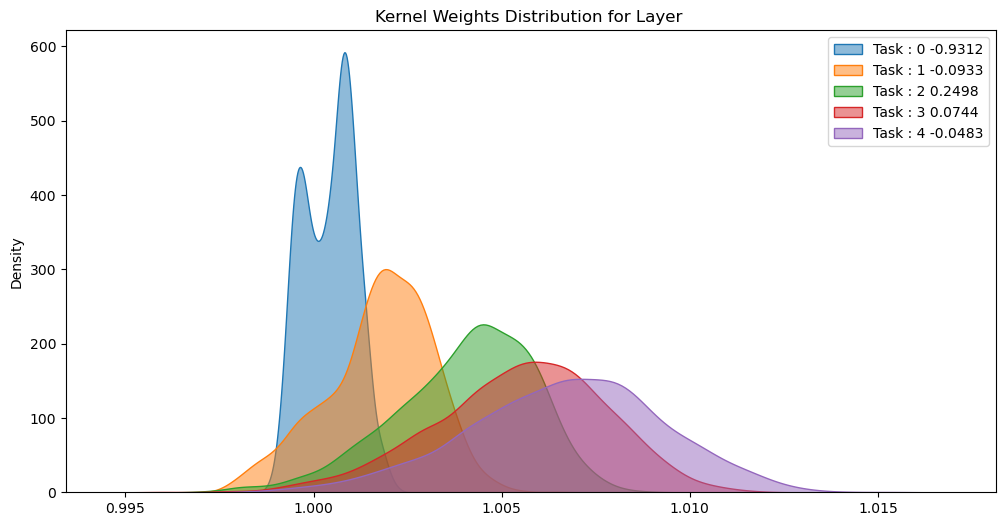}
		\caption{Kernel Density Estimation of DER + CB}
		\label{kdercb}
	\end{figure}
	
	\begin{figure}[H]
		\hspace{-0.3in}
		\includegraphics[scale=0.75]{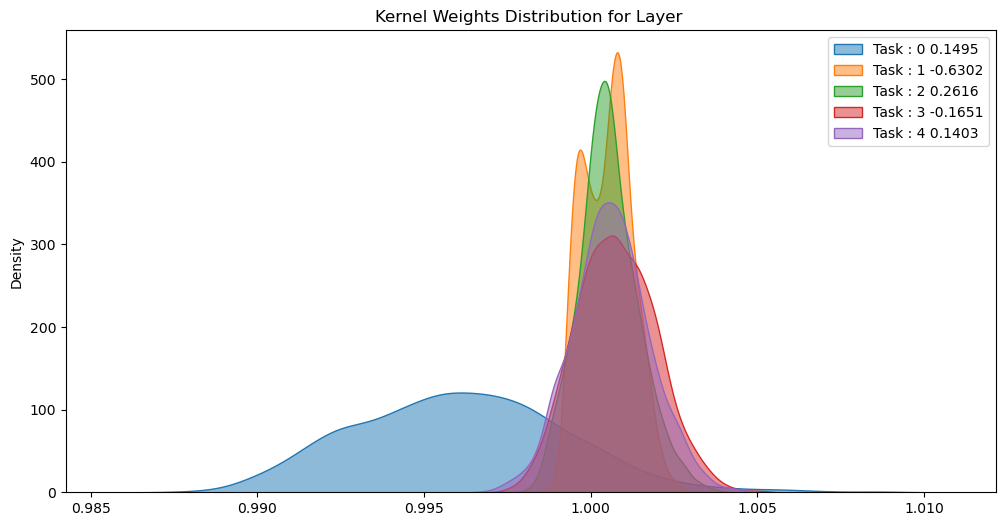}
		\caption{Kernel Density Estimation of FOSTER Teacher Network}
		\label{kfoster0}
	\end{figure}
	
	\begin{figure}[H]
		\hspace{-0.3in}
		\includegraphics[scale=0.75]{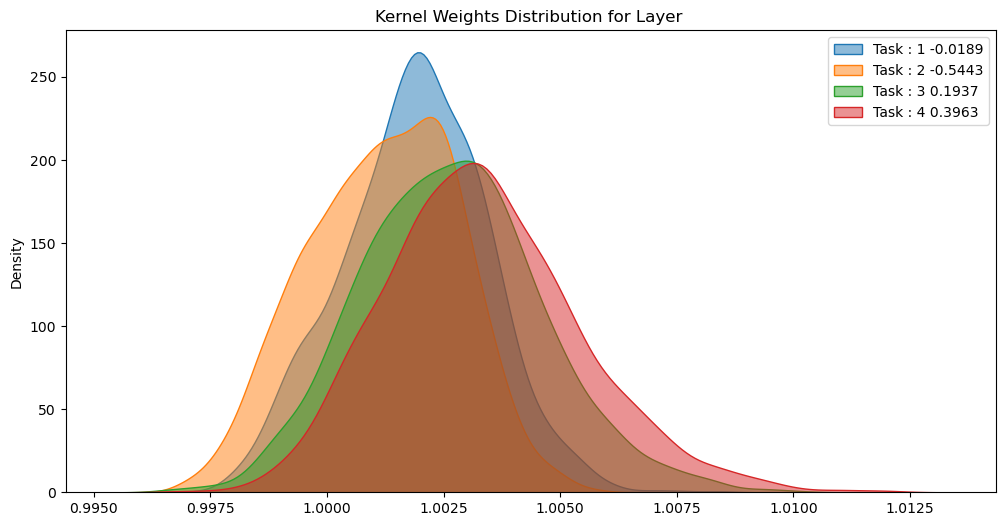}
		\caption{Kernel Density Estimation of FOSTER Student Network}
		\label{kfoster1}
	\end{figure}
	
	\begin{figure}[H]
		\hspace{-0.3in}
		\includegraphics[scale=0.75]{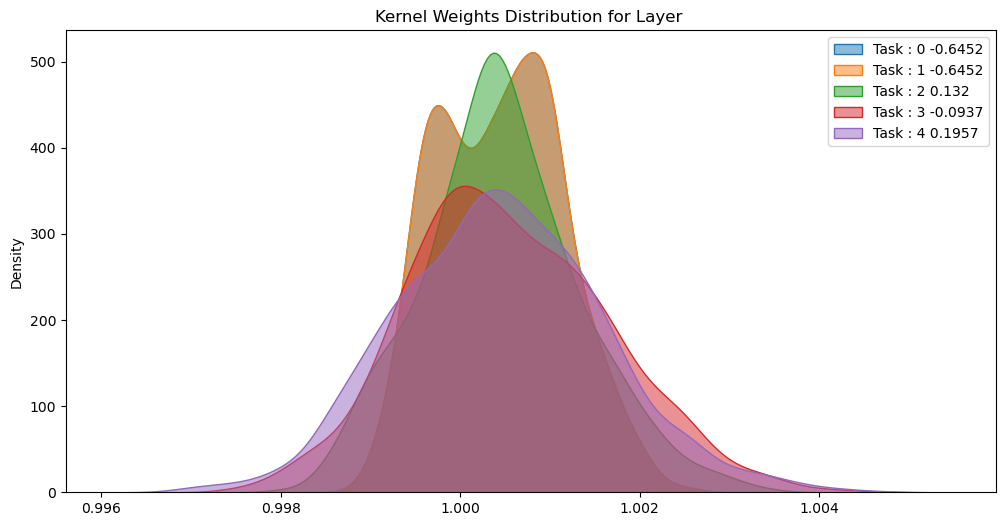}
		\caption{Kernel Density Estimation of FOSTER Teacher Network + CB}
		\label{kfoster0cb}
	\end{figure}
	
	\begin{figure}[H]
		\hspace{-0.3in}
		\includegraphics[scale=0.75]{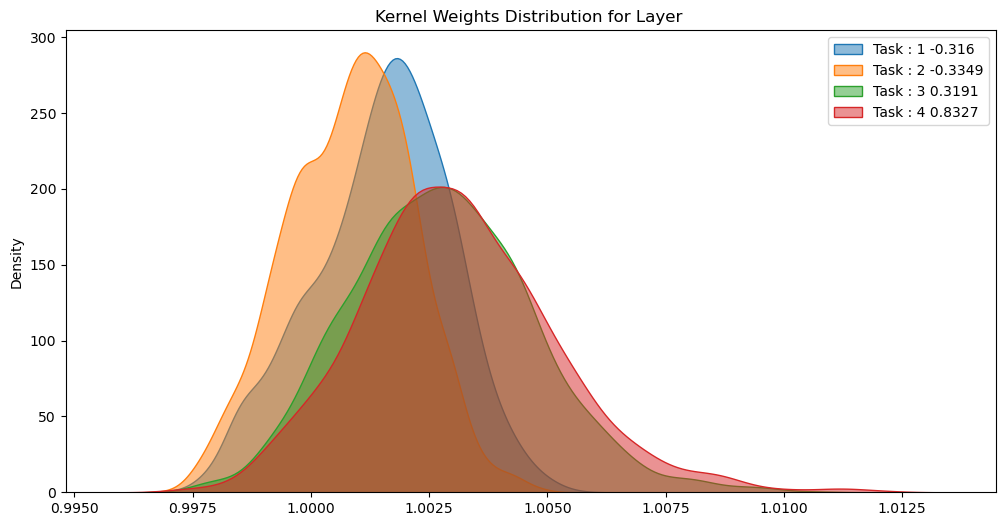}
		\caption{Kernel Density Estimation of FOSTER Student Network + CB}
		\label{kfoster1cb}
	\end{figure}
	
	\begin{figure}[H]
		\hspace{-0.3in}
		\includegraphics[scale=0.75]{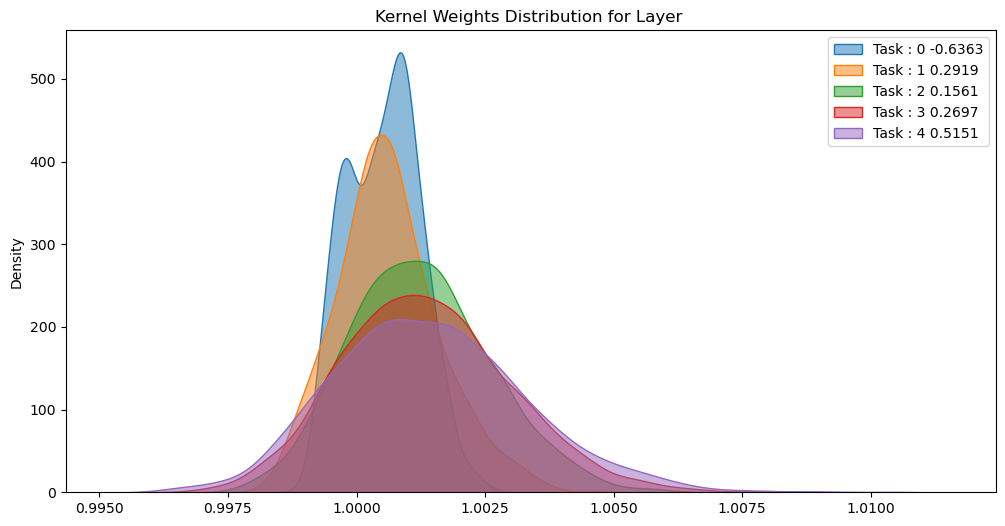}
		\caption{Kernel Density Estimation of MEMO}
		\label{kmemo0}
	\end{figure}
	
	\begin{figure}[H]
		\hspace{-0.3in}
		\includegraphics[scale=0.75]{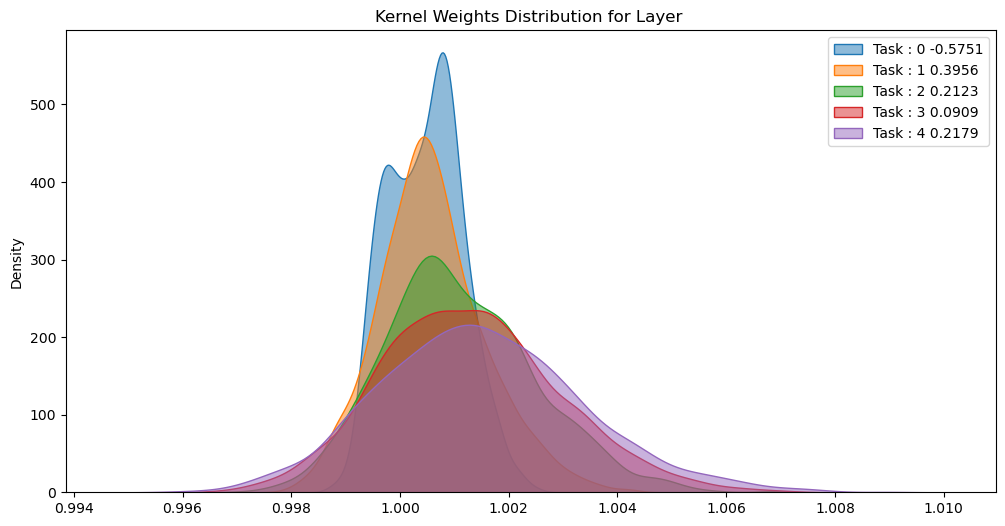}
		\caption{Kernel Density Estimation of MEMO + CB}
		\label{kmemo0cb}
	\end{figure}

\end{document}